\pdfoutput=1

\documentclass[11pt]{article}
\PassOptionsToPackage{dvipsnames, svgnames, table}{xcolor}
\usepackage[preprint]{acl}

\usepackage{times}
\usepackage{latexsym}

\usepackage[T1]{fontenc}

\usepackage[utf8]{inputenc}

\usepackage{microtype}

\usepackage{inconsolata}

\usepackage{graphicx}
\usepackage{booktabs}
\usepackage{makecell}
\usepackage{amsmath,amsfonts}

\usepackage{times}
\usepackage{latexsym}
\usepackage{float}
\usepackage{placeins}
\usepackage{mydef}

\usepackage[T1]{fontenc}      
\usepackage[utf8]{inputenc}   
\usepackage{microtype}        
\usepackage{inconsolata}      
\usepackage{parskip}          

\usepackage{colortbl}        
\usepackage{tcolorbox}       
\definecolor{mygray}{gray}{.88}
\definecolor{pyblue}{rgb}{0.0, 0.0, 0.5}
\definecolor{pygreen}{rgb}{0.0, 0.5, 0.0}
\definecolor{pyorange}{rgb}{1.0, 0.4, 0.0}
\definecolor{Gray}{gray}{0.9}

\usepackage{amsmath}
\usepackage{amssymb}
\usepackage{amsfonts}
\usepackage{amsthm}

\usepackage{graphicx}
\usepackage{subcaption}
\usepackage{adjustbox}
\usepackage{booktabs}       
\usepackage{tabularx}       
\usepackage{multirow}       
\usepackage{pifont}         
\usepackage{lscape}         
\usepackage{subfloat}

\usepackage{enumitem}
\usepackage{algorithm}
\usepackage{algpseudocode}

\usepackage{listings}
\lstdefinestyle{pythonstyle}{
    language=Python,
    basicstyle=\footnotesize\ttfamily,
    breaklines=true,
    morekeywords={self},
    keywordstyle=\color{pyblue},
    commentstyle=\color{pygreen},
    stringstyle=\color{pyorange},
    numberstyle=\tiny\color{gray},
    numbers=left,
    numbersep=10pt,
    tabsize=4,
    showspaces=false,
    showstringspaces=false
}

\usepackage{tikz}
\usetikzlibrary{shapes,positioning,trees,arrows,decorations.pathreplacing,shadows.blur}

\usepackage{url}

\usepackage{CJKutf8}        
\usepackage{framed}         
\usepackage{bbding}         
\usepackage{alltt}          
\usepackage{natbib}         
\usepackage{fontawesome5}   

\newcommand{\cmark}{\color{ForestGreen}\ding{51}}%
\newcommand{\xmark}{\color{Red}\ding{55}}%

\newlength\savewidth

\newtheorem{problem}{Problem Definition}
\title{A Survey of Large Language Models for Text-Guided Molecular Discovery: from Molecule Generation to Optimization}

\author{
  \textbf{Ziqing Wang\textsuperscript{1}\thanks{Equal Contribution}} \quad
  \textbf{Kexin Zhang\textsuperscript{1}\footnotemark[1]} \quad
  \textbf{Zihan Zhao\textsuperscript{1}} \quad
  \textbf{Yibo Wen\textsuperscript{1}} \\
  \textbf{Abhishek Pandey\textsuperscript{2}} \quad
  \textbf{Han Liu\textsuperscript{1}} \quad
  \textbf{Kaize Ding\textsuperscript{1}\thanks{Corresponding Author}} \\
  \textsuperscript{1}Northwestern University \quad
  \textsuperscript{2}AbbVie \\
  \texttt{\{ziqingwang2029, zihanzhao2026, yibowen2024\}@u.northwestern.edu} \\
  \texttt{kevin.kxzhang@gmail.com} \quad
  \texttt{abhishek.pandey@abbvie.com} \\
  \texttt{\{hanliu, kaize.ding\}@northwestern.edu}
}

\begin{document}
\maketitle
\begin{abstract}



Large language models (LLMs) are introducing a paradigm shift in molecular discovery by enabling text-guided interaction with chemical spaces through natural language, symbolic notations, with emerging extensions to incorporate multi-modal inputs. To advance the new field of LLM for molecular discovery, this survey provides an up-to-date and forward-looking review of the emerging use of LLMs for two central tasks: molecule generation and molecule optimization. Based on our proposed taxonomy for both problems, we analyze representative techniques in each category, highlighting how LLM capabilities are leveraged across different learning settings. In addition, we include the commonly used datasets and evaluation protocols. We conclude by discussing key challenges and future directions, positioning this survey as a resource for researchers working at the intersection of LLMs and molecular science. A continuously updated reading list is available at \url{https://github.com/REAL-Lab-NU/Awesome-LLM-Centric-Molecular-Discovery}.

\end{abstract}

\section{Introduction}
\label{sec:introduction}

Molecular design and optimization are fundamental to multiple scientific disciplines, including drug discovery~\citep{zheng2024large}, materials science~\citep{grandi2025evaluating}, and synthetic chemistry~\citep{lu2024generative,wang2025efficient}. 
However, these tasks present significant challenges due to the vast and complex chemical spaces that must be navigated to discover novel compounds with desirable properties while maintaining chemical validity and structural plausibility~\citep{zheng2024large,yu2025collaborative}. 
Over the years, a range of computational approaches has been developed to achieve these goals, from Variational Autoencoders~\citep{gomez2018automatic} and Generative Adversarial Networks~\citep{de2018molgan} to Transformers~\citep{edwards2022translation}. 
However, these traditional methods often struggle with generating high-quality, diverse, and synthesizable molecules~\citep{ramos2025review, sun2025synllama}.

More recently, large language models (LLMs) have emerged as particularly powerful tools for tackling these challenges, drawing increasing research attention~\citep{zheng2024large}. 
These foundation models, characterized by billions of parameters, exhibit emergent capabilities such as advanced reasoning, instruction following, and in-context learning, enabled by extensive pre-training on diverse datasets~\citep{brown2020language, wei2022emergent}. 
Thus, LLMs can leverage their extensive pre-training knowledge to generalize across chemical problems and can be further adapted to specialized tasks through fine-tuning. 
These unique capabilities have established LLMs as a powerful new paradigm for exploring chemical space and accelerating molecular discovery.

Despite the growing interest in applying LLMs to molecular discovery tasks, existing literature reviews fail to provide a comprehensive analysis of this specific intersection. 
Most earlier surveys~\citep{cheng2021molecular,zeng2022deep,tang2024survey,yang2024molecule} focus broadly on general deep generative AI approaches rather than specifically examining LLMs' unique contributions. 
Other reviews that do mention LLMs~\citep{ramos2025review, zhang2025scientific, guo2025survey, abunasser2024large, janakarajan2024language,liao2024words} either primarily focus on the general chemical domain or include smaller language models lacking the emergent capabilities characteristic of the LLMs central to this survey.

Our survey addresses this critical gap by providing the first overview specifically focused on LLMs as generators in molecular discovery, with particular emphasis on two central tasks: \textbf{molecule generation} and \textbf{molecule optimization}. 
Our survey specifically highlights how LLMs are deployed, adapted, and trained for navigating and manipulating complex chemical spaces,  distinguishing their role from auxiliary functions like feature extraction~\citep{liu2023multi} or control~\citep{liu2024multimodal}. 
Unlike prior surveys that categorize studies based on model architectures~\citep{abunasser2024large, janakarajan2024language}, we introduce a new taxonomy centered on the learning paradigms employed to leverage LLMs for generative molecular tasks. 
As illustrated in Fig.~\ref{fig:taxonomy}, we distinguish between approaches that operate \textit{without LLM tuning} (i.e., \textit{Zero-Shot Prompting} and \textit{In-Context Learning}) and those \textit{with LLM tuning} (i.e., \textit{Supervised Fine-Tuning} and \textit{Preference Tuning}), allowing researchers to better understand the effectiveness and limitations of different LLM utilization strategies.

To summarize, we provide the first systematic review focused on LLMs for text-guided molecular discovery for both generation and optimization tasks. The main contributions are as follows:

\begin{itemize}[leftmargin=*, itemsep=0pt, topsep=0.1pt]
\item  We introduce a new taxonomy categorizing existing research based on learning paradigms, revealing how different approaches utilize LLMs' capabilities, alongside their respective advantages and limitations.
\item  We provide a systematic summary of commonly used datasets, benchmarks, and evaluation metrics, offering a comprehensive reference for researchers in the field.
\item  We identify critical challenges and outline promising future research directions to further advance this rapidly evolving domain of LLM-centric molecular discovery.
\end{itemize}

\definecolor{darkgray}{gray}{0.7} 
\tikzstyle{leaf}=[draw=black,
    line width=0.4pt,
    rounded corners,minimum height=1em,
    fill=mygreen,text opacity=1, align=center,
    fill opacity=.5,  text=black,align=left,font=\scriptsize,
    inner xsep=3pt,
    inner ysep=1pt,
    ]
\tikzstyle{leaf_v2}=[draw=black,
    line width=0.4pt,
    rounded corners,minimum height=1em,
    fill=mypink,text opacity=1, align=center,
    fill opacity=.5,  text=black,align=left,font=\scriptsize,
    inner xsep=3pt,
    inner ysep=1pt,
    ]
\tikzstyle{middle}=[draw=black,
    line width=0.4pt,
    rounded corners,minimum height=1em,
    fill=myyellow,text opacity=1, align=center,
    fill opacity=.5,  text=black,align=center,font=\scriptsize,
    inner xsep=3pt,
    inner ysep=1pt,
    ]
\tikzstyle{middle_v2}=[draw=black,
    line width=0.4pt,
    rounded corners,minimum height=1em,
    fill=myblue,text opacity=1, align=center,
    fill opacity=.5,  text=black,align=center,font=\scriptsize,
    inner xsep=3pt,
    inner ysep=1pt,
    ]

\begin{figure*}[ht]
\vspace{-0.4cm}
\centering
\begin{forest}
  for tree={
  forked edges,
  grow=east,
  reversed=true,
  anchor=base west,
  parent anchor=east,
  child anchor=west,
  base=middle,
  font=\scriptsize,
  rectangle,
  line width=0.7pt,
  draw=output-black,
  rounded corners,align=left,
  minimum width=2em,
    s sep=5pt,
    inner xsep=3pt,
    inner ysep=1pt,
  },
  where level=1{text width=4.5em}{},
  where level=2{text width=6em,font=\scriptsize}{}, 
  where level=3{font=\scriptsize}{},
  where level=4{font=\scriptsize}{},
  where level=5{font=\scriptsize}{},
  [LLM-Centric Molecular Discovery, middle,rotate=90,anchor=north,edge=output-black
  [Generation, middle, edge=output-black, text width=3.5em
    [w/o Tuning, middle, edge=output-black, text width=3.5em 
            [In-Context Learning, middle_v2, text width=6.5em, edge=output-black 
                [LLM4GraphGen~\citep{yao2024exploring}{,} MolReGPT~\citep{li2024empowering}{,} \\ FrontierX~\citep{srinivas2024cross},leaf, text width=19.5em, edge=output-black] 
            ]
    ] 
    [w/ Tuning, middle, edge=output-black, text width=3.5em 
            [Supervised Fine-Tuning, middle_v2, text width=6.5em, edge=output-black 
                [Mol-instructions~\citep{fang2023mol}{,} LlaSMol~\citep{yu2024llasmol}{,} \\
                ChemLLM~\citep{zhang2024chemllm}{,} ICMA~\citep{li2024large}{,} \\
                MolReFlect~\citep{li2024molreflect}{,} ChatMol~\citep{fan2025chatmol}{,} \\
                PEIT-LLM~\citep{lin2025prop}{,} NatureLM~\citep{xia2025naturelm}{,} \\ SynLlama~\citep{sun2025synllama}{,} TOMG-Bench~\citep{li2024tomg}{,} \\
                UniMoT~\citep{zhang2024unimot}, leaf, text width=19.5em, edge=output-black] 
            ]
            [Preference Tuning, middle_v2, text width=6.5em, edge=output-black 
                [Div-SFT~\citep{jang2024can}{,} Mol-MOE~\citep{calanzone2025mol}{,} \\SmileyLLama~\citep{cavanagh2024smileyllama}{,} ALMol~\citep{gkoumas2024almol}{,} \\ Less for More~\citep{gkoumas2024less}{,} Mol-LLM~\citep{lee2025mol}, leaf, text width=19.5em, edge=output-black] 
    ]] 
]
[Optimization, middle, edge=output-black, text width=3.5em
    [w/o Tuning, middle, edge=output-black, text width=3.5em 
            [Zero-Shot Prompting, middle_v2, text width=6.5em, edge=output-black 
                [LLM-MDE~\citep{bhattacharya2024large}{,} MOLLEO~\citep{wang2025efficient},leaf_v2, text width=19.5em, edge=output-black] 
            ]
            [In-Context Learning, middle_v2, text width=6.5em, edge=output-black 
                [CIDD~\citep{gao2025pushing}{,} LLM-EO~\citep{lu2024generative}{,}\\ MOLLM~\citep{ran2025multi}{,} ChatDrug~\citep{liu2024conversational}{,}\\ Re$^2$DF~\citep{le2024utilizing}{,}  BOPRO~\citep{agarwalsearching},leaf_v2, text width=19.5em, edge=output-black] 
            ]
    ] 
    [w/ Tuning, middle, edge=output-black, text width=3.5em 
        [Supervised Fine-Tuning, middle_v2, text width=6.5em, edge=output-black 
            [MultiMol~\citep{yu2025collaborative}{,} DrugAssist~\citep{ye2025drugassist}{,} \\ GeLLM$^3$O~\citep{dey2025mathtt}{,} DrugLLM~\citep{liu2024drugllm}{,} \\ LLM-Enhanced GA~\citep{bedrosian2024small}{,} \\ Molx-Enhanced LLM~\citep{le2024molx}{,} TOMG-Bench~\citep{li2024tomg}, leaf_v2, text width=19.5em, edge=output-black] 
        ]
        [Preference Tuning, middle_v2, text width=6.5em, edge=output-black 
                [NatureLM~\citep{xia2025naturelm},leaf_v2, text width=19.5em, edge=output-black] 
            ]
    ] 
]
]] 
\end{forest} 
\caption{A Taxonomy of LLM-Centric Molecular Discovery.}
\label{fig:taxonomy}
\vspace{-0.1cm}
\end{figure*}
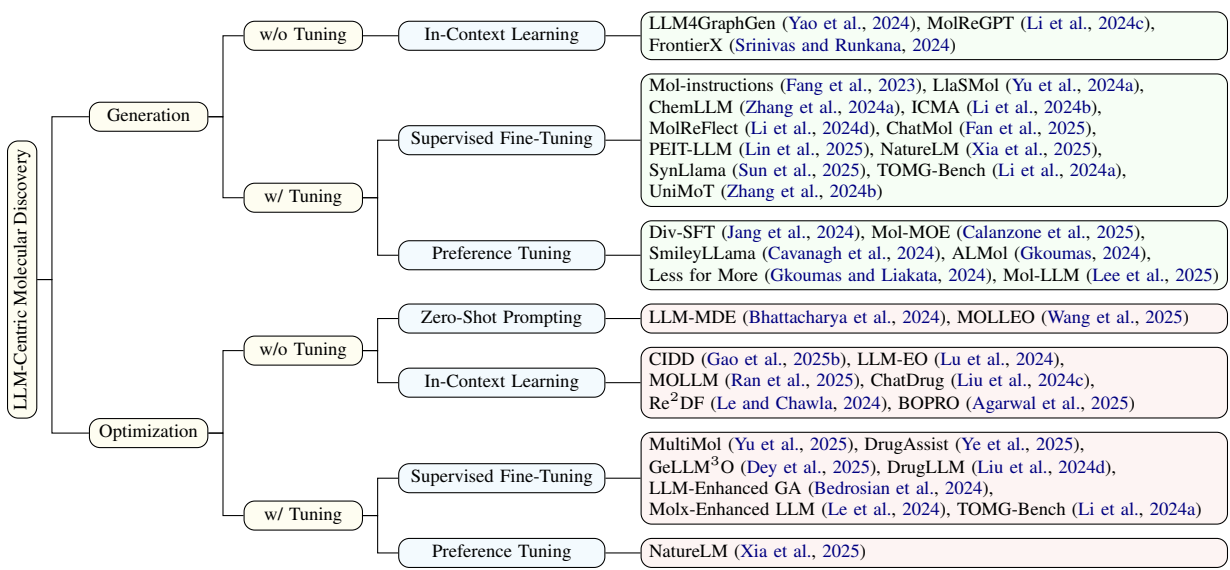



\section{Preliminaries}
\label{sec:Pre}
\subsection{Large Language Models} 
\label{subsec:Def_LLM}
LLMs distinguish themselves from earlier Pre-trained Language Models (PLMs) like BERT~\citep{devlin2019bert} (which typically possessed millions of parameters) primarily through their massive scale, often boasting parameter counts in the billions, and the resultant emergent capabilities not found in smaller models~\citep{zhao2023survey, yang2023harnessing}. The development of these LLMs and their advanced functionalities is largely attributed to their pre-training on vast text corpora, predominantly through an autoregressive next-token prediction objective. This immense scale facilitates emergent capabilities~\citep{wei2022emergent} such as in-context learning~\citep{brown2020language}, chain-of-thought reasoning~\citep{wei2022chain}, and powerful zero-shot generalization, which are not consistently observed in their smaller predecessors. These advanced capabilities render LLMs uniquely suited for tackling complex chemical applications like molecule generation and optimization tasks central to this review. \textbf{For clarity and scope within this survey, we focus specifically on foundation models with at least 1 billion (1B) parameters}.

\subsection{Problem Definition}
In this survey, we focus on two central tasks:
\vspace{0.2cm}
\begin{problem}[\textbf{LLM-centric Molecule Generation}] 
\label{prob:molecule_generation}
This task leverages LLMs as the core generative engine for the de novo design of novel molecular structures based on specified input instructions.
\end{problem}

\vspace{0.2cm}
\begin{problem}[\textbf{LLM-centric Molecule Optimization}]
\label{prob:molecule_optimization}
This task leverages LLMs to modify or edit a given input molecule, aiming to enhance one or more of its properties while often preserving essential structural characteristics.
\end{problem}

As illustrated in Fig.~\ref{fig: overview}, for both tasks, the input prompt provided to the LLM typically comprises three key components: 
(1) \textbf{\textit{Instruction ($\mathcal{I}$)}}: A textual component that defines the primary guidance and objectives of the task.
(2) \textbf{\textit{Few-Shot Examples ($E_{fs}$)}} (Optional): A small set of input-output examples relevant to the task, provided to facilitate in-context learning.
(3) \textbf{\textit{Property Constraints ($\mathcal{C}_{p}$)}} (Optional): Explicit desired values, ranges, or thresholds for specific molecular properties.

While these input components are common to both tasks, their specific content and function differ significantly. \textbf{For Molecule Generation}, the Instruction $\mathcal{I}$ typically consists of a natural language description of the desired molecular characteristics or a general task definition. The objective is to generate a chemically valid molecular representation (e.g., a SMILES string $S_M$) that aligns with this instruction and any provided property constraints $\mathcal{C}_p$, potentially guided by few-shot examples $E_{fs}$. \textbf{For Molecule Optimization}, the Instruction $\mathcal{I}$ serves a more specific purpose. It not only outlines the optimization objectives but also crucially includes an \textbf{initial molecule $M_x$} that requires modification. This initial molecule can be represented in various formats, such as a 1D sequence (e.g., SMILES), a 2D graph, or 3D coordinates (see Appendix~\ref{appendix:data_modality} for details). The instruction typically specifies which properties should be improved. The objective is to generate a chemically valid modified molecule (e.g., $S_{M_y}$) that enhances the desired properties of $M_x$ while adhering to any specified constraints $\mathcal{C}_p$, potentially guided by few-shot examples $E_{fs}$.

\subsection{Learning Paradigms}

\label{subsec:learning_paradigms}

\begin{figure*}[t!]  
\vspace{-0.6cm}
\centering  
\includegraphics[height=6.2cm]{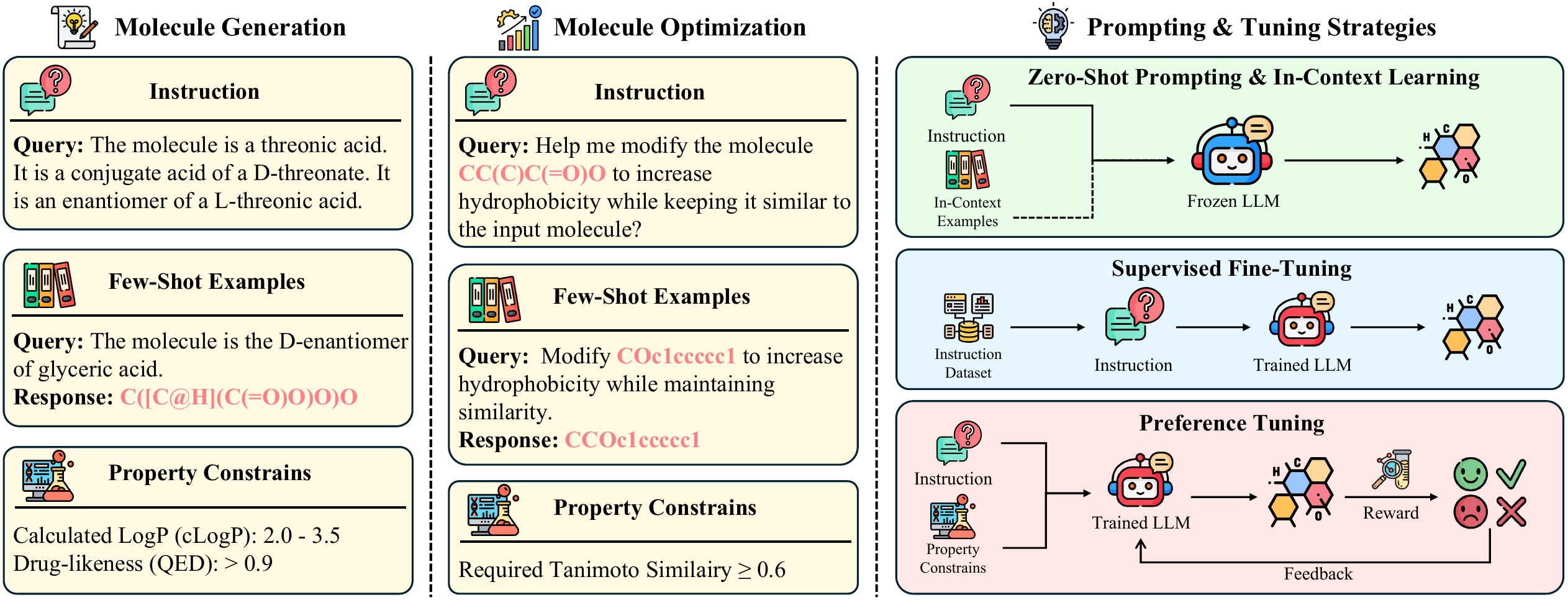}
\caption{\textbf{Overview of LLM-Centric Molecular Discovery.} \textbf{\textit{Left:}} Typical input components (Instruction, Few-Shot Examples, Property Constraints) for molecule generation and optimization. \textbf{\textit{Right:}} Core learning paradigms for applying LLMs to \textit{Zero-Shot Prompting \& In-Context Learning}, \textit{Supervised Fine-Tuning} and \textit{Preference Tuning}.}
\label{fig: overview}  
\vspace{-0.1cm}
\end{figure*}

The application of LLMs to molecular discovery tasks, as depicted in the taxonomy in Fig.~\ref{fig: overview}, can be broadly categorized based on whether the model's parameters are updated for the specific task. This distinction defines two primary learning paradigms:

\textbf{Without LLM Tuning:} These methods utilize pre-trained LLMs directly, guiding their behavior solely through the input prompt~$\mathcal{I}$ without modifying the model's weights. This paradigm primarily encompasses strategies like \textit{Zero-Shot Prompting}, where the LLM operates based on instructions alone, and \textit{In-Context Learning (ICL)}, where few-shot examples provided within the prompt guide the model's responses. These approaches avoid computationally training but rely heavily on the LLM's inherent capabilities and effective prompt engineering.

\textbf{With LLM Tuning:} These methods involve adapting the pre-trained LLM by further training and updating its parameters to specialize it for molecular tasks or align its outputs with desired objectives. This typically includes \textit{Supervised Fine-Tuning (SFT)}, where the model learns from labeled task-specific datasets, and subsequent \textit{Preference Tuning} (or Alignment), where the model is refined based on feedback. While tuning can significantly enhance performance, it requires curated data and computational resources.

\section{Molecule Generation}
\label{sec:generation}
Molecule generation, the computational creation of novel molecular structures, is a cornerstone of modern drug discovery and materials science~\citep{elton2019deep}. This section reviews recent advances in LLM-centric molecule generation, primarily categorizing approaches based on the learning paradigms defined in Section~\ref{subsec:learning_paradigms}.


\subsection{Molecule Generation without Tuning}

\textbf{In-Context Learning:} Since \textit{Zero-Shot Prompting} is challenging for general-purpose LLMs due to their lack of specialized chemical knowledge, most successful applications in this paradigm heavily rely on ICL to provide specific guidance. For instance, FrontierX~\citep{srinivas2024cross} uses knowledge-augmented prompting, supplying detailed instructions alongside few-shot examples within the prompt to guide de novo design effectively. Similarly, LLM4GraphGen~\citep{yao2024exploring} explores property-based generation by prompting LLMs with target properties and relevant molecular examples, evaluating performance under different prompting strategies, including few-shot ICL. Recognizing the importance of example quality, MolReGPT~\citep{li2024empowering} incorporates Retrieval-Augmented Generation (RAG), dynamically retrieving highly relevant molecule-caption pairs to serve as more effective few-shot context, thereby boosting ICL performance.


\subsection{Molecule Generation with Tuning}

\textbf{Supervised Fine-Tuning:} While non-tuning methods leverage pre-trained knowledge effectively, their capabilities can be limited for highly specialized or complex generation tasks. 
SFT addresses this by adapting pre-trained LLMs specifically for molecule generation on labeled datasets, typically pairs of instructions and target molecular representations. 
Although early explorations demonstrated the viability of SFT using smaller PLMs such as MolGPT~\citep{bagal2021molgpt} and MolT5~\citep{edwards2022translation}, current research focuses on harnessing large foundation models.

A predominant SFT strategy is the \textbf{curation of large-scale, high-quality instruction datasets} to instill chemical knowledge into general-purpose LLMs (as shown in Fig.~\ref{fig: cap2mol}). 
Initiatives such as LlaSMol~\citep{yu2024llasmol} with its SMolInstruct dataset, ChemLLM~\citep{zhang2024chemllm} with ChemData, Mol-Instructions~\citep{fang2023mol} covering broader biomolecular text, and the OpenMolIns dataset from TOMG-Bench~\citep{li2024tomg} all exemplify this trend. 
These efforts fine-tune models like LLaMA-2-7B~\citep{touvron2023llama} and Mistral-7B~\citep{jiang2023mistral} with LoRA~\citep{hu2021lora}, to enhance instruction following and performance on the molecule generation task.

Beyond broad instruction tuning, SFT methodologies also address specific challenges in molecule generation. A significant hurdle is ensuring that generated molecules precisely meet complex constraints. ChatMol~\citep{fan2025chatmol} directly addresses this limitation by using a numerical enhancement technique, significantly improving the model's fidelity to specified quantitative property values. Concurrently, SynLlama~\citep{sun2025synllama} tackles synthetic feasibility to generate complete synthetic pathways. Other innovative SFT strategies include integrating dynamic context directly into the fine-tuning process; ICMA~\citep{li2024large} and MolReFlect~\citep{li2024molreflect} propose In-Context Molecule Tuning (ICMT), which fine-tunes the LLM with relevant retrieved examples. Furthermore, PEIT-LLM~\citep{lin2025prop} proposes a two-step Property Enhanced Instruction Tuning (PEIT) framework, first synthesizing instruction data with a multi-modal model, then using it to fine-tune LLMs for tasks like multi-constraint generation. NatureLM~\citep{xia2025naturelm} demonstrates the application of SFT on models pre-trained across multiple scientific domains for tasks including text-instructed molecule generation.

However, the SFT methods discussed above primarily operate on text-based representations (like SMILES or SELFIES), which inherently struggle to explicitly encode rich structural information crucial for chemistry. \textbf{Multi-modal SFT} approaches aim to bridge this gap by incorporating these richer data types. UniMoT~\citep{zhang2024unimot} exemplifies a solution by introducing a novel molecule tokenizer. Leveraging Vector Quantization (VQ) and a Causal Q-Former, this component converts graph-based molecular features into discrete "molecule tokens", enabling unified autoregressive processing of text and graph-derived molecular information.


\textbf{Preference Tuning:} Following SFT, which primarily teaches models to mimic static input-output patterns from datasets, \textit{Preference Tuning} techniques offer further refinement by employing feedback-driven learning to shape LLM outputs towards desired characteristics.
In molecule generation, this feedback is typically incorporated in two main ways: (1) \textbf{RL-based methods}~\citep{sutton1998reinforcement} optimize the LLM (policy) using a scalar reward signal derived from evaluating generated molecules against desired criteria. 
(2) \textbf{Offline methods} like Direct Preference Optimization (DPO) learn from preference pairs ("chosen" vs "rejected") of molecules, training the LLM to assign higher likelihoods to the preferred candidates based on comparative evaluations.

SmileyLlama~\citep{cavanagh2024smileyllama} utilizes DPO after SFT to significantly improve adherence to specified property constraints by learning from preferences between correctly and incorrectly generated molecules. 
Mol-MoE~\citep{calanzone2025mol} uses a preference objective to train a Mixture-of-Experts router for molecule generation. 
Furthermore, Div-SFT~\citep{jang2024can}, after an initial SFT stage, employs RL with a reward function explicitly designed to maximize structural diversity among the generated molecules. 
Similarly, contrastive methods like Contrastive Preference Optimization (CPO)~\cite{xu2024contrastive} have been used to refine the quality and relevance of generated molecules based on preference data comparing desired targets against less optimal alternatives, proving effective even with limited data~\citep{gkoumas2024almol, gkoumas2024less}.

\textit{Preference Tuning} is not limited to text-only input but also can handle \textbf{multi-modal} inputs after SFT. 
These approaches focus on improving how the model utilizes structural information, although this remains a more nascent area of research. 
For example, Mol-LLM~\citep{lee2025mol} demonstrates better leveraging of 2D graph inputs through Molecular Structure Preference Optimization (MolPO). 
After an initial SFT phase involving graph inputs, MolPO further trains the LLM using preference pairs where the distinction between "chosen" and "rejected" outputs is based on the correctness of the input molecular graph conditioning the generation. 
This preference learning implicitly guides the model to better integrate and leverage the provided structural information during processing.

\section{Molecule Optimization}
\label{sec:optimization}
Molecule optimization is the task of refining molecular structures to improve one or more desired properties, such as solubility, binding affinity, or synthetic accessibility. Unlike molecule generation, optimization starts with an initial molecule and proposes targeted structural modifications to achieve specific goals. 
This section summarizes LLM-centric molecule optimization methods, with a focus on how different learning paradigms (see Section~\ref{subsec:learning_paradigms}) are leveraged to guide optimization.

\subsection{Molecule Optimization without Tuning}

\textbf{Zero-Shot Prompting:}
\textit{Zero-Shot Prompting} leverages the pre-trained capabilities of LLMs to modify input molecules according to natural language instructions, without providing specific examples in the prompt. This setting assumes that the model can interpret molecular structure (often via SMILES) and property-related text well enough to perform molecule optimization.
For example,
LLM-MDE~\citep{bhattacharya2024large} guides optimization with natural language prompts that specify desired property changes and structural constraints, enabling controlled modifications to given parent molecules.
MOLLEO, on the other hand, integrates LLMs into an evolutionary framework inspired by population-based algorithms~\citep{jensen2019graph}.
It uses prompt-based sampling to generate candidates through mutations and crossovers, while applying filtering steps to enforce structural similarity.
These methods demonstrate the flexibility of zero-shot prompting in expressing diverse optimization goals, though they often struggle with precise control in multi-objective settings.

\textbf{In-Context Learning:}
In contrast, ICL incorporates examples of previous molecular edits into the prompt. This allows the LLM to learn optimization strategies by modifying new molecules in ways consistent with observed property improvements or structural changes.
CIDD~\citep{gao2025pushing} structures molecule optimization into a multi-step pipeline: interaction analysis, design, and reflection. Each step is guided by prompts derived from interaction profiles, and during the design step, previous designs and reflections are provided to make better modifications.  

Both LLM-EO~\citep{lu2024generative} and MOLLM~\citep{ran2025multi} integrate LLMs into an \textbf{Evolutionary Algorithm} (EA) framework through in-context prompting.  
LLM-EO specifically targets transition metal complexes and guides optimization through prompts that include both objectives and examples of successful or failed complexes, enabling iterative improvement across generations.  
MOLLM eliminates external operators entirely, using the LLM to perform all genetic operations. The model is guided by structured prompt templates containing optimization goals, molecular context, and historical experience. It includes modules for candidate selection (via Pareto and scalarized scoring), and prompt construction, all designed for effective in-context molecule refinement.

Retrieval-augmented prompting further strengthens ICL by retrieving structurally similar and high-performing molecules from a given database.
ChatDrug~\citep{liu2024conversational} retrieves structurally similar molecules and incorporates this information into the prompt context, allowing the LLM to iteratively propose refinements based on feedback. 
Re$^2$DF~\citep{le2024utilizing} enhances this paradigm by integrating chemical validity feedback via RDKit~\citep{landrum2013rdkit}. When invalid molecules are generated, the resulting error messages are used as feedback, closing the loop and guiding the LLM toward valid outputs.
Additionally, recent work by BOPRO~\citep{agarwalsearching} combines ICL with Bayesian optimization. 
A surrogate model scores generated candidates and proposes updated prompts that include high-quality examples from the search history. 
The LLM then uses these prompts to generate new SMILES strings, forming a feedback-driven, example-conditioned optimization cycle.

\subsection{Molecule Optimization with Tuning}

\textbf{Supervised Fine-Tuning:}
SFT adapts pre-trained LLMs to molecule optimization tasks by training on curated datasets that pair molecular inputs with corresponding optimized outputs, under explicit property-based instructions. 
These datasets often include transformation examples where the input molecule is associated with property modification goals (e.g., improving solubility or binding affinity) and the corresponding optimized molecules.
Through such supervision, the model learns how to perform controlled structural edits conditioned on specific objectives.

\tikzstyle{leaf}=[draw=black,
    line width=0.4pt,
    rounded corners,minimum height=1em,
    fill=mygreen,text opacity=1, align=center,
    fill opacity=.5,  text=black,align=left,font=\scriptsize,
    inner xsep=3pt,
    inner ysep=1pt,
    ]
\tikzstyle{leaf_v2}=[draw=black,
    line width=0.4pt,
    rounded corners,minimum height=1em,
    fill=mypink,text opacity=1, align=center,
    fill opacity=.5,  text=black,align=left,font=\scriptsize,
    inner xsep=3pt,
    inner ysep=1pt,
    ]
\tikzstyle{middle}=[draw=black,
    line width=0.4pt,
    rounded corners,minimum height=1em,
    fill=myyellow,text opacity=1, align=center,
    fill opacity=.5,  text=black,align=center,font=\scriptsize,
    inner xsep=3pt,
    inner ysep=1pt,
    ]
\tikzstyle{middle_v2}=[draw=black,
    line width=0.4pt,
    rounded corners,minimum height=1em,
    fill=myblue,text opacity=1, align=center,
    fill opacity=.5,  text=black,align=center,font=\scriptsize,
    inner xsep=3pt,
    inner ysep=1pt,
    ]

\begin{figure*}[ht]
\vspace{-4mm}
\centering
\begin{forest}
  for tree={
    forked edges,
    grow=east,
    reversed=true,
    anchor=base west,
    parent anchor=east,
    child anchor=west,
    base=middle,
    font=\scriptsize,
    rectangle,
    line width=0.7pt,
    draw=output-black,
    rounded corners,
    align=left,
    minimum width=2em,
    s sep=5pt,
    inner xsep=3pt,
    inner ysep=1pt,
  },
  where level=1{text width=4.5em}{},
  where level=2{text width=6em,font=\scriptsize}{},
  where level=3{font=\scriptsize}{},
  where level=4{font=\scriptsize}{},
  where level=5{font=\scriptsize}{},
  [Benchmarking \& Evaluation, middle,rotate=90,anchor=north,edge=output-black
    [Datasets, middle, edge=output-black, text width=2.5em
      [Pretraining-Only, middle, edge=output-black, text width=7em
    [ZINC~\citep{irwin2012zinc}{,} PubChem~\citep{kim2016pubchem, kim2019pubchem, kim2025pubchem}{,} \\
    ChemData~\citep{zhang2024chemllm}{,} MuMOInstruct~\citep{dey2025mathtt}{,}\\Mol-Instructions~\citep{fang2023mol}, leaf, text width=25.2em, edge=output-black]
  ]
  [Benchmark-Only, middle, edge=output-black, text width=7em
    [MoleculeNet~\citep{wu2018moleculenet}{,} ChemBench~\citep{mirza2404large}{,} \\ MOSES~\citep{polykovskiy2020molecular}{,} TOMG-Bench~\citep{li2024tomg} , leaf, text width=25.2em, edge=output-black]
  ]
  [Pretraining \& Benchmark, middle, edge=output-black, text width=7em
    [ChEMBL~\citep{gaulton2012chembl}{,} ChEBI-20~\citep{edwards2021text2mol}{,} QM9~\citep{pinheiro2020machine}{,} \\CrossDocked2020~\citep{francoeur2020three}{,} Dockstring~\citep{garcia2022dockstring}{,} \\ MolOpt-Instructions~\citep{ye2025drugassist}{,} L+M-24~\citep{edwards2024l+}{,} \\ SMolInstruct~\citep{yu2024advancing}{,} OGBG-MolHIV~\citep{hu2020open}, leaf, text width=25.2em, edge=output-black]
  ]]
    [Metrics, middle, edge=output-black, text width=2.5em
      [Structure-Based, middle, edge=output-black, text width=7.0em
[Validity \& Similarity, middle_v2, text width=6.3em, edge=output-black, 
  [Validity~\citep{polykovskiy2020molecular}{,} EM~\citep{rajpurkar2016squad}{,}\\
   BLEU~\citep{papineni2002bleu}{,} Levenshtein~\citep{levenshtein1966binary}{,} \\
   FTS(MACCS ~\citep{durant2002reoptimization}{,} RDK~\citep{landrum2013rdkit}{,}\\Morgan~\citep{morgan1965generation}){,} FCD~\citep{preuer2018frechet},
   leaf_v2, text width=17.2em, edge=output-black
  ]
]
        [Diversity \& Uniqueness, middle_v2, text width=6.3em, edge=output-black
          [NCircle~\citep{jang2024can}{,}IntDiv~\citep{benhenda2017chemgan}{,}\\Novel Rate~\citep{brown2019guacamol}{,} Unique@1k~\citep{wang2023cmol}{,} \\
Unique@10k~\citep{bagal2021molgpt},leaf_v2, text width=17.2em, edge=output-black]
        ]
      ]
      [Property-Based, middle, edge=output-black, text width=7em
  [Single-Property, middle_v2, edge=output-black, text width=6.3em
    [LogP~\citep{hansch1968linear}{,} TPSA~\citep{ertl2000fast}{,} \\SA score~\citep{ertl2009estimation} \\
    QED~\citep{bickerton2012quantifying}, leaf_v2, text width=17.2em, edge=output-black]
  ]
  [Multi-Property, middle_v2, edge=output-black, text width=6.3em
    [Success Rate under Constraints~\citep{jin2020multi}{,}\\ Pareto Optimality~\citep{pareto1919manuale}{,}\\
    Composite Score~\citep{jin2020multi}, leaf_v2, text width=17.2em, edge=output-black]
  ]
]
]
  ]]]
\end{forest}
\vspace{-5mm}
\caption{A Taxonomy of Benchmarking \& Evaluation in Molecule Discovery.}
\vspace{-3mm}
\label{fig:evaluation_taxonomy}
\end{figure*}
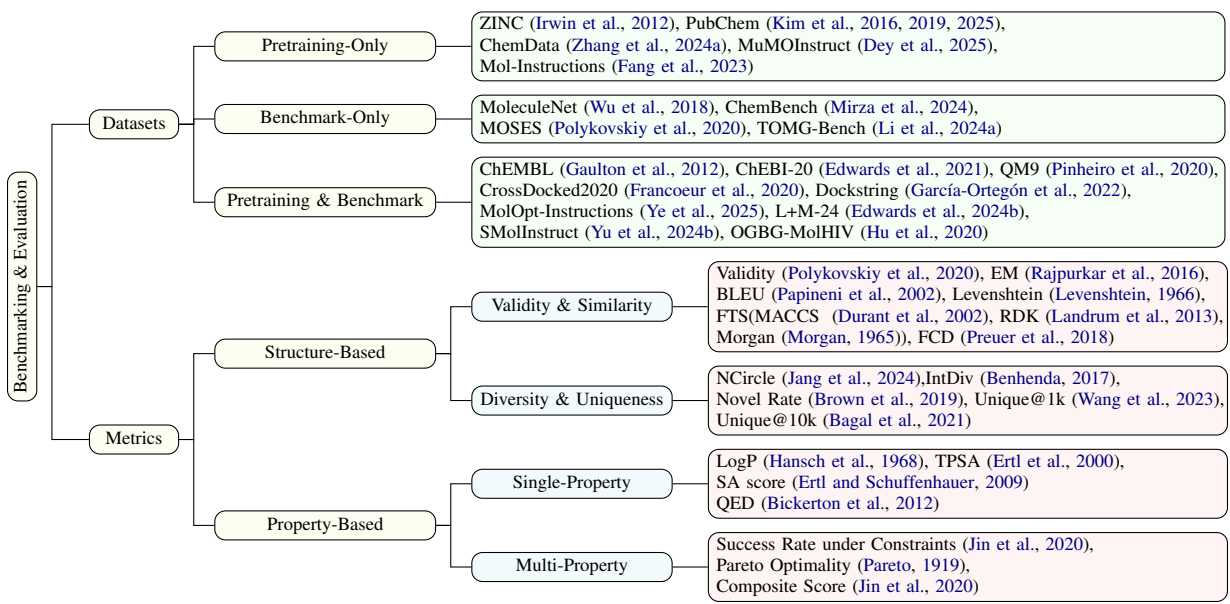

Several recent methods leverage SFT to improve the ability of LLMs to conduct molecule edits.
DrugAssist~\citep{ye2025drugassist} fine-tunes LLaMA-2-7B-Chat using a curated instruction dataset MolOpt-Instructions and adopts a multi-task learning strategy that combines general conversational data and molecule-specific instructions, helping preserve interactivity while learning task-specific patterns.  
However, its focus on single- and dual-property tasks limits scalability to more complex objectives.
To address this limitation, GeLLM$^3$O~\citep{dey2025mathtt} proposes an instruction-tuned framework for multi-property optimization. It introduces MuMOInstruct, a dataset curated for diverse objectives, and trains both specialist and generalist models. The generalist variant shows strong generalization to novel out-of-distribution tasks without retraining, demonstrating potential for instruction-tuned LLMs as flexible optimization engines.
In addition, MultiMol~\citep{yu2025collaborative} represents a collaborative framework combining a fine-tuned worker model and a research agent. The worker, trained on over one million molecules, reconstructs SMILES based on scaffold-property prompts and modulates them during inference for property optimization. The research agent (GPT-4o) extracts structure–property patterns from literature and ranks candidates using regression-based scoring, ensuring consistency with domain-specific knowledge.

Transformer-based chemical language models (CLM)~\citep{ross2022large,ross2024learning,wu2024leveraging,dai2025zero,liu2025controllablegpt} have demonstrated strong potential for molecule optimization tasks. 
Unlike prior models that rely on raw SMILES sequences, DrugLLM~\citep{liu2024drugllm} introduces a group-based molecular representation (GMR) that encodes SMILES strings to align structure and semantics. It adopts an autoregressive training objective to model the generative process of molecular modifications conditioned on property descriptions or prior examples.

SFT also plays a key role in population-based optimization frameworks.
LLM-Enhanced GA~\citep{bedrosian2024small} 
proposes an iterative process in which new candidates are generated via prompt-based sampling from high-performing molecules, replacing traditional mutation and crossover.
Explicit oracle modeling is incorporated through supervised fine-tuning on evaluated molecules when performance stagnates, allowing the LLM to progressively refine its understanding of structure–property relationships.

Beyond text-only molecule optimization, \textbf{multi-modal molecule optimization} incorporates structural information such as molecular graphs and 3D geometries. These additional modalities enable more accurate modeling of structure–property relationships and improve control over chemical validity~\citep{zhang2024deep,lin2024versatile,nakamura2025molecular}. 
Molx-Enhanced LLM~\citep{le2024molx} exemplifies this approach with a framework that integrates SMILES strings, 2D molecular graphs, and handcrafted fingerprints into a unified embedding. It employs LLaMA-2-7B as the base LLM and introduces a trainable multi-modal module, MolX, which is pre-trained with supervised molecule–text pairs and auxiliary tasks to align molecular representations with the LLM's textual input space. Importantly, during fine-tuning, the use of graph encoders and fingerprint integration ensures that the model captures both global topology and substructural details, which are essential for chemically valid optimization. It indicates that fine-tuning the LLM to establish multi-modal models shows better performances than generalist chemical LLMs.

\textbf{Preference Tuning:}
\textit{Preference Tuning} aims to adjust large language models to better follow human instructions, preferences, or task-specific goals~\citep{park2025mol,chen2025generalists}. In molecule optimization, alignment techniques help models generate molecules that meet specific optimization criteria more reliably.
RL-based alignment techniques, such as DrugImproverGPT~\citep{liu2025drugimprovergpt} and ScaffoldGPT~\citep{liu2025scaffoldgpt}, both built on Transformer-based architectures, explicitly incorporate reward functions to guide molecule optimization.
Moving beyond reliance on direct supervised signals, NatureLM~\citep{xia2025naturelm} augments its post-trained 8B model using DPO, to improve molecule optimization across nine pharmacologically relevant properties. 
Instead of training on absolute labels or scalar rewards, the model is optimized using a curated dataset of 179.5k prompt–response preference pairs, where each instance presents a "preferred" and a "rejected" molecular output given the same prompt.
By training with DPO, NatureLM demonstrates improved alignment with desirable molecular properties and generalizes preference-guided optimization across diverse chemical objectives.

\section{Benchmarking and Evaluation}
\label{sec:benchmarking_evaluation}

Rigorous benchmarking and comprehensive evaluation are crucial for tracking the progress of LLM-centric molecular discovery. This section provides an overview of the common resources and methodologies used, focusing on the datasets that form the basis of benchmarking efforts and the metrics applied for robust evaluation. Our discussion is structured around the taxonomy presented in Fig.~\ref{fig:evaluation_taxonomy}.

\subsection{Datasets}
\label{subsec:datasets_benchmark}
A variety of datasets serve as the foundation for training and benchmarking LLMs in molecular discovery, differing in their primary utility:
\textbf{Pretraining-Only Datasets} provide vast quantities of unlabeled molecular structures or general chemical knowledge, such as ZINC~\citep{irwin2012zinc} and PubChem~\citep{kim2025pubchem}, or large-scale instruction collections like ChemData~\citep{zhang2024chemllm}.
\textbf{Benchmark-Only Datasets} are smaller, curated collections designed for specific evaluation tasks. Examples include TOMG-Bench~\citep{li2024tomg} for open-domain molecule generation, and MOSES~\citep{polykovskiy2020molecular} for de novo design benchmarking.
A third category comprises datasets suitable for both \textbf{Pre-training and Benchmark} applications, offering a balance of scale and task-specificity. Notable examples include ChEMBL~\citep{gaulton2012chembl} for bioactivity data, and instruction datasets like SMolInstruct~\citep{yu2024advancing}.
Further details and a comparative summary of these and other relevant datasets are available in Appendix~\ref{appendix: datasets} and Table~\ref{tab:datasets_summary}.

\subsection{Metrics}
\label{subsec:metrics_evaluation}

The performance of LLMs in molecular tasks is critically assessed using a diverse set of metrics, broadly categorized into structure-based and property-based evaluations, which are essential for quantifying success in both molecule generation and optimization.
\textbf{Structure-Based Metrics} evaluate the intrinsic quality and diversity of molecular structures, including (1) \textit{Validity and Similarity} metrics, which assess chemical correctness and resemblance to reference structures (e.g., validity rate, exact match); and (2) \textit{Diversity and Uniqueness} metrics, which quantify the variety and novelty of the generated outputs (e.g., uniqueness rate, novelty rate). \textbf{Property-Based Metrics} gauge how well molecules meet desired functional criteria, applied for (1) \textit{Single-Property} evaluation focusing on individual targets like Quantitative Estimate of Drug-likeness (QED), Lipophilicity (LogP), Synthetic Accessibility (SA), and binding affinity; and (2) \textit{Multi-Property} evaluation assessing performance across several objectives, often via composite scores or success rates under multiple constraints. A comprehensive catalogue and detailed discussion of these evaluation metrics are provided in Appendix~\ref{appendix:metrics}.

\section{Conclusion and Future Work}
\label{sec:conclusion}

This survey presents the first comprehensive review of recent advances in LLM-centric molecular discovery, covering both generation and optimization.  
We introduce a novel taxonomy distinguishing approaches based on different learning paradigms—specifically, without LLM tuning (e.g., zero-shot prompting, in-context learning) versus with LLM tuning (e.g., supervised fine-tuning, preference tuning).
This framework allows for a systematic analysis of how current strategies leverage LLM capabilities, revealing key trends, strengths, and limitations.
The rapid progress in this field underscores LLMs' transformative potential to accelerate scientific discovery in chemistry and related disciplines. However, several challenges and exciting opportunities remain for future research:

\textbf{Trustworthy Generation and Hallucination Mitigation:} While LLMs can generate chemically plausible molecules, they often produce outputs that are chemically invalid or factually incorrect without domain-specific supervision~\citep{le2024utilizing}. This lack of transparency limits their applicability in high-stakes domains such as drug development~\citep{ma2025reasoning}. While interpretable prompting and rationalization techniques~\citep{xiao2025m} offer promising solutions, controlled hallucinations may actually serve as a creativity mechanism, potentially uncovering novel molecular scaffolds inaccessible through conventional search methods~\citep{edwards2024language,yuan2025hallucinations}. The future challenge lies not in eliminating hallucinations entirely, but in developing frameworks that can distinguish between harmful fabrications and beneficial creative leaps.

\textbf{LLM Agents for Interactive Discovery:} LLMs are increasingly being integrated into agent-based frameworks, where they coordinate with external tools (e.g., retrosynthesis engines, docking software, or lab automation platforms) to complete multi-step discovery workflows~\citep{feng2025retool,liu2025advances}. Building robust LLM agents that can plan, reason, and interact with both humans and tools could enable more flexible and goal-directed molecular design~\citep{gao2025pharmagents}. These agents could potentially close the loop between computational prediction and experimental validation, accelerating the iterative discovery process.

\textbf{Multi-Modal Modeling and Alignment:} Incorporating multiple molecular modalities remains a core challenge. Most current LLM-based approaches typically treat modalities separately, with limited cross-modal interaction. Future work should prioritize architectures that unify these representations, allowing joint encoding and reasoning over chemical topology, geometry, and textual semantics~\citep{lu2023graphgpt,pirnay2025graphxform}. By developing sophisticated tokenization and fusion techniques that bridge discrete and continuous representations, future systems could achieve a more holistic understanding of chemical structures and properties, potentially leading to more accurate and innovative molecular designs.

\section*{Limitations}
\label{sec:limitations}
This survey focuses on the use of large language models for two core tasks in text-guided molecular discovery: molecule generation and molecule optimization. These tasks represent the most direct applications of LLMs in molecular design and are the primary scope of current research.
We are aware that LLMs can also significantly impact other important areas of molecular science~\citep{sun2025synllama}, such as reaction prediction, retrosynthesis, protein–ligand modeling, and automated experimentation~\citep{zhang2024comprehensive,liu2024scientific,liu2025quantitative}. Given the broad and rapidly evolving landscape, we leave a systematic review of these directions to future work.
By narrowing the scope of this work, we provide a focused and detailed resource for researchers working on LLM-driven molecular design
In the future, we anticipate expanding this analysis to encompass these additional domains as the field continues to evolve.



\bibliography{reference}

\begin{thebibliography}{123}
\providecommand{\natexlab}[1]{#1}

\bibitem[{AbuNasser(2024)}]{abunasser2024large}
Raghad AbuNasser. 2024.
\newblock \href {https://chemrxiv.org/engage/api-gateway/chemrxiv/assets/orp/resource/item/66fbff36cec5d6c142ca3852/original/large-language-models-in-drug-discovery-a-survey.pdf} {Large language models in drug discovery: A survey}.

\bibitem[{Agarwal et~al.(2025)Agarwal, Arivazhagan, Das, Swamy, Khosla, and Gangadharaiah}]{agarwalsearching}
Dhruv Agarwal, Manoj~Ghuhan Arivazhagan, Rajarshi Das, Sandesh Swamy, Sopan Khosla, and Rashmi Gangadharaiah. 2025.
\newblock \href {https://openreview.net/pdf?id=aVfDrl7xDV} {Searching for optimal solutions with llms via bayesian optimization}.
\newblock In \emph{The Thirteenth International Conference on Learning Representations}.

\bibitem[{Bagal et~al.(2021)Bagal, Aggarwal, Vinod, and Priyakumar}]{bagal2021molgpt}
Viraj Bagal, Rishal Aggarwal, PK~Vinod, and U~Deva Priyakumar. 2021.
\newblock \href {https://chemrxiv.org/engage/api-gateway/chemrxiv/assets/orp/resource/item/60c7588e469df48597f456ae/original/lig-gpt-molecular-generation-using-a-transformer-decoder-model.pdf} {Molgpt: molecular generation using a transformer-decoder model}.
\newblock \emph{Journal of chemical information and modeling}, 62(9):2064--2076.

\bibitem[{Bedrosian et~al.(2024)Bedrosian, Guevorguian, Fahradyan, Chilingaryan, Khachatrian, and Aghajanyan}]{bedrosian2024small}
Menua Bedrosian, Philipp Guevorguian, Tigran Fahradyan, Gayane Chilingaryan, Hrant Khachatrian, and Armen Aghajanyan. 2024.
\newblock \href {https://openreview.net/pdf?id=nJCYKdRZXb} {Small molecule optimization with large language models}.
\newblock In \emph{Neurips 2024 Workshop Foundation Models for Science: Progress, Opportunities, and Challenges}.

\bibitem[{Benhenda(2017)}]{benhenda2017chemgan}
Mostapha Benhenda. 2017.
\newblock \href {https://arxiv.org/pdf/1708.08227} {Chemgan challenge for drug discovery: can ai reproduce natural chemical diversity?}
\newblock \emph{arXiv preprint arXiv:1708.08227}.

\bibitem[{Bhattacharya et~al.(2024)Bhattacharya, Cassady, Hickner, and Reinhart}]{bhattacharya2024large}
Debjyoti Bhattacharya, Harrison~J Cassady, Michael~A Hickner, and Wesley~F Reinhart. 2024.
\newblock \href {https://chemrxiv.org/engage/api-gateway/chemrxiv/assets/orp/resource/item/66cf24e6f3f4b052906147bc/original/large-language-models-as-molecular-design-engines.pdf} {Large language models as molecular design engines}.
\newblock \emph{Journal of Chemical Information and Modeling}, 64(18):7086--7096.

\bibitem[{Bickerton et~al.(2012)Bickerton, Paolini, Besnard, Muresan, and Hopkins}]{bickerton2012quantifying}
GR~Bickerton, GV~Paolini, J~Besnard, S~Muresan, and AL~Hopkins. 2012.
\newblock \href {https://www.nature.com/articles/nchem.1243.pdf} {Quantifying the chemical beauty of drugs}.
\newblock \emph{Nature Chemistry}, 4(2):90--98.

\bibitem[{Brown et~al.(2019)Brown, Fiscato, Segler, and Vaucher}]{brown2019guacamol}
Nathan Brown, Marco Fiscato, Marwin~HS Segler, and Alain~C Vaucher. 2019.
\newblock \href {https://pubs.acs.org/doi/pdf/10.1021/acs.jcim.8b00839} {Guacamol: Benchmarking models for de novo molecular design}.
\newblock \emph{Journal of chemical information and modeling}, 59(3):1096--1108.

\bibitem[{Brown et~al.(2020)Brown, Mann, Ryder, Subbiah, Kaplan, Dhariwal, Neelakantan, Shyam, Sastry, Askell et~al.}]{brown2020language}
Tom Brown, Benjamin Mann, Nick Ryder, Melanie Subbiah, Jared~D Kaplan, Prafulla Dhariwal, Arvind Neelakantan, Pranav Shyam, Girish Sastry, Amanda Askell, and 1 others. 2020.
\newblock \href {https://proceedings.neurips.cc/paper_files/paper/2020/file/1457c0d6bfcb4967418bfb8ac142f64a-Paper.pdf} {Language models are few-shot learners}.
\newblock \emph{NeurIPS}, 33:1877--1901.

\bibitem[{Calanzone et~al.(2025)Calanzone, D'Oro, and Bacon}]{calanzone2025mol}
Diego Calanzone, Pierluca D'Oro, and Pierre-Luc Bacon. 2025.
\newblock \href {https://arxiv.org/pdf/2502.05633} {Mol-moe: Training preference-guided routers for molecule generation}.
\newblock \emph{arXiv preprint arXiv:2502.05633}.

\bibitem[{Cavanagh et~al.(2024)Cavanagh, Sun, Gritsevskiy, Bagni, Bannister, and Head-Gordon}]{cavanagh2024smileyllama}
Joseph~M Cavanagh, Kunyang Sun, Andrew Gritsevskiy, Dorian Bagni, Thomas~D Bannister, and Teresa Head-Gordon. 2024.
\newblock \href {https://arxiv.org/pdf/2409.02231} {Smileyllama: Modifying large language models for directed chemical space exploration}.
\newblock \emph{arXiv preprint arXiv:2409.02231}.

\bibitem[{Chen et~al.(2025)Chen, Stanton, Ding, Alberstein, Watkins, Bonneau, Gligorijević, Cho, and Frey}]{chen2025generalists}
Angelica Chen, Samuel~D. Stanton, Frances Ding, Robert~G. Alberstein, Andrew~M. Watkins, Richard Bonneau, Vladimir Gligorijević, Kyunghyun Cho, and Nathan~C. Frey. 2025.
\newblock \href {https://arxiv.org/pdf/2410.22296v4} {Generalists vs. specialists: Evaluating llms on highly-constrained biophysical sequence optimization tasks}.

\bibitem[{Cheng et~al.(2021)Cheng, Gong, Liu, Song, and Zou}]{cheng2021molecular}
Yu~Cheng, Yongshun Gong, Yuansheng Liu, Bosheng Song, and Quan Zou. 2021.
\newblock \href {https://drive.google.com/file/d/10W-VoybuxHqFDYsUyfaJWfx-zuh6cBIA/view} {Molecular design in drug discovery: a comprehensive review of deep generative models}.
\newblock \emph{Briefings in bioinformatics}, 22(6):bbab344.

\bibitem[{Dai et~al.(2025)Dai, Zhang, Zhong, Fu, Deng, Zhang, Liu, and Gao}]{dai2025zero}
Zhilian Dai, Jie Zhang, Songyou Zhong, Jiawei Fu, Yangyang Deng, Dan Zhang, Yichao Liu, and Peng Gao. 2025.
\newblock \href {https://chemrxiv.org/engage/api-gateway/chemrxiv/assets/orp/resource/item/6775f7e96dde43c908089ffa/original/a-zero-shot-single-point-molecule-optimization-model-mimicking-medicinal-chemists-expertise.pdf} {A zero-shot single-point molecule optimization model: Mimicking medicinal chemists’ expertise}.

\bibitem[{De~Cao and Kipf(2018)}]{de2018molgan}
Nicola De~Cao and Thomas Kipf. 2018.
\newblock \href {https://arxiv.org/pdf/1805.11973} {Molgan: An implicit generative model for small molecular graphs}.
\newblock \emph{arXiv preprint arXiv:1805.11973}.

\bibitem[{Devlin et~al.(2019)Devlin, Chang, Lee, and Toutanova}]{devlin2019bert}
Jacob Devlin, Ming-Wei Chang, Kenton Lee, and Kristina Toutanova. 2019.
\newblock \href {https://aclanthology.org/N19-1423.pdf} {Bert: Pre-training of deep bidirectional transformers for language understanding}.
\newblock In \emph{Proceedings of the 2019 conference of the North American chapter of the association for computational linguistics: human language technologies, volume 1 (long and short papers)}, pages 4171--4186.

\bibitem[{Dey et~al.(2025)Dey, Hu, and Ning}]{dey2025mathtt}
Vishal Dey, Xiao Hu, and Xia Ning. 2025.
\newblock \href {https://arxiv.org/abs/2502.13398} {Gellm\({}^{\mbox{3}}\)o: Generalizing large language models for multi-property molecule optimization}.
\newblock \emph{arXiv preprint arXiv:2502.13398}.

\bibitem[{Durant et~al.(2002)Durant, Leland, Henry, and Nourse}]{durant2002reoptimization}
JL~Durant, BA~Leland, DR~Henry, and JG~Nourse. 2002.
\newblock \href {https://pubs.acs.org/doi/pdf/10.1021/ci010132r} {Reoptimization of mdl keys for use in drug discovery}.
\newblock \emph{Journal of Chemical Information and Computer Sciences}, 42(6):1273--1280.

\bibitem[{Edwards et~al.(2022)Edwards, Lai, Ros, Honke, Cho, and Ji}]{edwards2022translation}
Carl Edwards, Tuan Lai, Kevin Ros, Garrett Honke, Kyunghyun Cho, and Heng Ji. 2022.
\newblock \href {https://arxiv.org/pdf/2204.11817} {Translation between molecules and natural language}.
\newblock \emph{arXiv preprint arXiv:2204.11817}.

\bibitem[{Edwards et~al.(2024{\natexlab{a}})Edwards, Wang, and Ji}]{edwards2024language}
Carl Edwards, Qingyun Wang, and Heng Ji. 2024{\natexlab{a}}.
\newblock \href {https://aclanthology.org/2024.eacl-tutorials.3.pdf} {Language+ molecules}.
\newblock In \emph{Proceedings of the 18th Conference of the European Chapter of the Association for Computational Linguistics: Tutorial Abstracts}, pages 14--20.

\bibitem[{Edwards et~al.(2024{\natexlab{b}})Edwards, Wang, Zhao, and Ji}]{edwards2024l+}
Carl Edwards, Qingyun Wang, Lawrence Zhao, and Heng Ji. 2024{\natexlab{b}}.
\newblock \href {https://arxiv.org/pdf/2403.00791} {L+ m-24: Building a dataset for language+ molecules@ acl 2024}.
\newblock \emph{arXiv preprint arXiv:2403.00791}.

\bibitem[{Edwards et~al.(2021)Edwards, Zhai, and Ji}]{edwards2021text2mol}
Carl Edwards, ChengXiang Zhai, and Heng Ji. 2021.
\newblock \href {https://aclanthology.org/2021.emnlp-main.47.pdf} {Text2mol: Cross-modal molecule retrieval with natural language queries}.
\newblock In \emph{EMNLP}, pages 595--607.

\bibitem[{Elton et~al.(2019)Elton, Boukouvalas, Fuge, and Chung}]{elton2019deep}
Daniel~C Elton, Zois Boukouvalas, Mark~D Fuge, and Peter~W Chung. 2019.
\newblock \href {https://pubs.rsc.org/en/content/articlepdf/2019/me/c9me00039a} {Deep learning for molecular design—a review of the state of the art}.
\newblock \emph{Molecular Systems Design \& Engineering}, 4(4):828--849.

\bibitem[{Ertl et~al.(2000)Ertl, Rohde, and Selzer}]{ertl2000fast}
Peter Ertl, Bernhard Rohde, and Paul Selzer. 2000.
\newblock \href {https://pubs.acs.org/doi/pdf/10.1021/jm000942e?casa_token=Xba52t19k6gAAAAA:lRhh2ZvvUnCjjYFQ9W4Lv_WxVAQi8j8uXamgyjSonH-soDpf8lnyoZZ8G77JeZBUypMYQbBfg7IANEzKSw} {Fast calculation of molecular polar surface area as a sum of fragment-based contributions and its application to the prediction of drug transport properties}.
\newblock \emph{Journal of Medicinal Chemistry}, 43(20):3714--3717.

\bibitem[{Ertl and Schuffenhauer(2009)}]{ertl2009estimation}
Peter Ertl and Ansgar Schuffenhauer. 2009.
\newblock \href {https://link.springer.com/content/pdf/10.1186/1758-2946-1-8.pdf} {Estimation of synthetic accessibility score of drug-like molecules based on molecular complexity and fragment contributions}.
\newblock \emph{Journal of Cheminformatics}, 1(1):8.

\bibitem[{Fan et~al.(2025)Fan, Cao, Ma, Yu, Peng, Zhang, Gao, and Fu}]{fan2025chatmol}
Chuanliu Fan, Ziqiang Cao, Zicheng Ma, Nan Yu, Yimin Peng, Jun Zhang, Yiqin Gao, and Guohong Fu. 2025.
\newblock \href {https://arxiv.org/pdf/2502.19794} {Chatmol: A versatile molecule designer based on the numerically enhanced large language model}.
\newblock \emph{arXiv preprint arXiv:2502.19794}.

\bibitem[{Fang et~al.(2023)Fang, Liang, Zhang, Liu, Huang, Chen, Fan, and Chen}]{fang2023mol}
Yin Fang, Xiaozhuan Liang, Ningyu Zhang, Kangwei Liu, Rui Huang, Zhuo Chen, Xiaohui Fan, and Huajun Chen. 2023.
\newblock \href {https://arxiv.org/pdf/2306.08018} {Mol-instructions: A large-scale biomolecular instruction dataset for large language models}.
\newblock \emph{arXiv preprint arXiv:2306.08018}.

\bibitem[{Favre and Powell(2014)}]{iupac2013nomenclature}
Henri~A Favre and Warren~H Powell. 2014.
\newblock \href {https://doi.org/10.1039/9781849733069} {\emph{Nomenclature of Organic Chemistry: IUPAC Recommendations and Preferred Names 2013}}.
\newblock Royal Society of Chemistry.

\bibitem[{Feng et~al.(2025)Feng, Huang, Qu, Zhang, Qin, Zhong, Jiang, Chi, and Zhong}]{feng2025retool}
Jiazhan Feng, Shijue Huang, Xingwei Qu, Ge~Zhang, Yujia Qin, Baoquan Zhong, Chengquan Jiang, Jinxin Chi, and Wanjun Zhong. 2025.
\newblock \href {https://arxiv.org/pdf/2504.11536?} {Retool: Reinforcement learning for strategic tool use in llms}.
\newblock \emph{arXiv preprint arXiv:2504.11536}.

\bibitem[{Francoeur et~al.(2020)Francoeur, Masuda, Sunseri, Jia, Iovanisci, Snyder, and Koes}]{francoeur2020three}
Paul~G Francoeur, Tomohide Masuda, Jocelyn Sunseri, Andrew Jia, Richard~B Iovanisci, Ian Snyder, and David~R Koes. 2020.
\newblock \href {https://pubs.acs.org/doi/pdf/10.1021/acs.jcim.0c00411} {Three-dimensional convolutional neural networks and a cross-docked data set for structure-based drug design}.
\newblock \emph{Journal of chemical information and modeling}, 60(9):4200--4215.

\bibitem[{Gao et~al.(2025{\natexlab{a}})Gao, Huang, Liu, Xie, Ma, Zhang, and Lan}]{gao2025pharmagents}
Bowen Gao, Yanwen Huang, Yiqiao Liu, Wenxuan Xie, Wei-Ying Ma, Ya-Qin Zhang, and Yanyan Lan. 2025{\natexlab{a}}.
\newblock \href {https://arxiv.org/pdf/2503.22164} {Pharmagents: Building a virtual pharma with large language model agents}.
\newblock \emph{arXiv preprint arXiv:2503.22164}.

\bibitem[{Gao et~al.(2025{\natexlab{b}})Gao, Huang, Liu, Xie, Ma, Zhang, and Lan}]{gao2025pushing}
Bowen Gao, Yanwen Huang, Yiqiao Liu, Wenxuan Xie, Wei-Ying Ma, Ya-Qin Zhang, and Yanyan Lan. 2025{\natexlab{b}}.
\newblock \href {https://arxiv.org/pdf/2503.01376} {Pushing the boundaries of structure-based drug design through collaboration with large language models}.
\newblock \emph{arXiv preprint arXiv:2503.01376}.

\bibitem[{Garc{\'\i}a-Orteg{\'o}n et~al.(2022)Garc{\'\i}a-Orteg{\'o}n, Simm, Tripp, Hern{\'a}ndez-Lobato, Bender, and Bacallado}]{garcia2022dockstring}
Miguel Garc{\'\i}a-Orteg{\'o}n, Gregor~NC Simm, Austin~J Tripp, Jos{\'e}~Miguel Hern{\'a}ndez-Lobato, Andreas Bender, and Sergio Bacallado. 2022.
\newblock \href {https://pubs.acs.org/doi/pdf/10.1021/acs.jcim.1c01334} {Dockstring: easy molecular docking yields better benchmarks for ligand design}.
\newblock \emph{Journal of chemical information and modeling}, 62(15):3486--3502.

\bibitem[{Gaulton et~al.(2012)Gaulton, Bellis, Bento, Chambers, Davies, Hersey, Light, McGlinchey, Michalovich, Al-Lazikani et~al.}]{gaulton2012chembl}
Anna Gaulton, Louisa~J Bellis, A~Patricia Bento, Jon Chambers, Mark Davies, Anne Hersey, Yvonne Light, Shaun McGlinchey, David Michalovich, Bissan Al-Lazikani, and 1 others. 2012.
\newblock \href {https://academic.oup.com/nar/article-pdf/40/D1/D1100/16955876/gkr777.pdf} {Chembl: a large-scale bioactivity database for drug discovery}.
\newblock \emph{Nucleic acids research}, 40(D1):D1100--D1107.

\bibitem[{Gkoumas(2024)}]{gkoumas2024almol}
Dimitris Gkoumas. 2024.
\newblock \href {https://arxiv.org/pdf/2405.08619} {Almol: Aligned language-molecule translation llms through offline preference contrastive optimisation}.
\newblock \emph{arXiv preprint arXiv:2405.08619}.

\bibitem[{Gkoumas and Liakata(2024)}]{gkoumas2024less}
Dimitris Gkoumas and Maria Liakata. 2024.
\newblock \href {https://arxiv.org/pdf/2405.13984} {Less for more: Enhanced feedback-aligned mixed llms for molecule caption generation and fine-grained nli evaluation}.
\newblock \emph{arXiv preprint arXiv:2405.13984}.

\bibitem[{G{\'o}mez-Bombarelli et~al.(2018)G{\'o}mez-Bombarelli, Wei, Duvenaud, Hern{\'a}ndez-Lobato, S{\'a}nchez-Lengeling, Sheberla, Aguilera-Iparraguirre, Hirzel, Adams, and Aspuru-Guzik}]{gomez2018automatic}
Rafael G{\'o}mez-Bombarelli, Jennifer~N Wei, David Duvenaud, Jos{\'e}~Miguel Hern{\'a}ndez-Lobato, Benjam{\'\i}n S{\'a}nchez-Lengeling, Dennis Sheberla, Jorge Aguilera-Iparraguirre, Timothy~D Hirzel, Ryan~P Adams, and Al{\'a}n Aspuru-Guzik. 2018.
\newblock \href {https://pubs.acs.org/doi/pdf/10.1021/acscentsci.7b00572} {Automatic chemical design using a data-driven continuous representation of molecules}.
\newblock \emph{ACS central science}, 4(2):268--276.

\bibitem[{Grandi et~al.(2025)Grandi, Jain, Groom, Cramer, and McComb}]{grandi2025evaluating}
Daniele Grandi, Yash~Patawari Jain, Allin Groom, Brandon Cramer, and Christopher McComb. 2025.
\newblock \href {https://asmedigitalcollection.asme.org/computingengineering/article-pdf/25/2/021004/7404714/jcise_25_2_021004.pdf} {Evaluating large language models for material selection}.
\newblock \emph{Journal of Computing and Information Science in Engineering}, 25(2):021004.

\bibitem[{Guo et~al.(2025)Guo, Xing, Zhou, Jiang, Chen, Wang, Jiang, Wang, Hou, Jiang et~al.}]{guo2025survey}
Huijie Guo, Xudong Xing, Yongjie Zhou, Wenjiao Jiang, Xiaoyi Chen, Ting Wang, Zixuan Jiang, Yibing Wang, Junyan Hou, Yukun Jiang, and 1 others. 2025.
\newblock \href {https://ieeexplore.ieee.org/iel8/6287639/6514899/10930479.pdf} {A survey of large language model for drug research and development}.
\newblock \emph{IEEE Access}.

\bibitem[{Hansch et~al.(1968)Hansch, Quinlan, and Lawrence}]{hansch1968linear}
Corwin Hansch, John~E Quinlan, and Gary~L Lawrence. 1968.
\newblock \href {https://pubs.acs.org/doi/pdf/10.1021/jo01265a071} {Linear free-energy relationship between partition coefficients and the aqueous solubility of organic liquids}.
\newblock \emph{The journal of organic chemistry}, 33(1):347--350.

\bibitem[{Hu et~al.(2021)Hu, Wallis, Allen-Zhu, Li, Wang, Wang, Chen et~al.}]{hu2021lora}
Edward~J Hu, Phillip Wallis, Zeyuan Allen-Zhu, Yuanzhi Li, Shean Wang, Lu~Wang, Weizhu Chen, and 1 others. 2021.
\newblock \href {https://arxiv.org/pdf/2106.09685v1/1000} {Lora: Low-rank adaptation of large language models}.
\newblock In \emph{ICLR}.

\bibitem[{Hu et~al.(2020)Hu, Fey, Zitnik, Dong, Ren, Liu, Catasta, and Leskovec}]{hu2020open}
Weihua Hu, Matthias Fey, Marinka Zitnik, Yuxiao Dong, Hongyu Ren, Bowen Liu, Michele Catasta, and Jure Leskovec. 2020.
\newblock \href {https://proceedings.neurips.cc/paper/2020/file/fb60d411a5c5b72b2e7d3527cfc84fd0-Paper.pdf} {Open graph benchmark: Datasets for machine learning on graphs}.
\newblock \emph{Advances in neural information processing systems}, 33:22118--22133.

\bibitem[{Irwin et~al.(2012)Irwin, Sterling, Mysinger, Bolstad, and Coleman}]{irwin2012zinc}
John~J Irwin, Teague Sterling, Michael~M Mysinger, Erin~S Bolstad, and Ryan~G Coleman. 2012.
\newblock \href {https://pubs.acs.org/doi/pdf/10.1021/ci3001277} {Zinc: a free tool to discover chemistry for biology}.
\newblock \emph{Journal of chemical information and modeling}, 52(7):1757--1768.

\bibitem[{Janakarajan et~al.(2024)Janakarajan, Erdmann, Swaminathan, Laino, and Born}]{janakarajan2024language}
Nikita Janakarajan, Tim Erdmann, Sarath Swaminathan, Teodoro Laino, and Jannis Born. 2024.
\newblock \href {https://arxiv.org/pdf/2309.16235} {Language models in molecular discovery}.
\newblock In \emph{Drug Development Supported by Informatics}, pages 121--141. Springer.

\bibitem[{Jang et~al.(2024)Jang, Jang, Kim, and Ahn}]{jang2024can}
Hyosoon Jang, Yunhui Jang, Jaehyung Kim, and Sungsoo Ahn. 2024.
\newblock \href {https://arxiv.org/pdf/2410.03138} {Can llms generate diverse molecules? towards alignment with structural diversity}.
\newblock \emph{arXiv preprint arXiv:2410.03138}.

\bibitem[{Jensen(2019)}]{jensen2019graph}
Jan~H Jensen. 2019.
\newblock \href {https://pubs.rsc.org/en/content/articlepdf/2013/91/c8sc05372c} {A graph-based genetic algorithm and generative model/monte carlo tree search for the exploration of chemical space}.
\newblock \emph{Chemical science}, 10(12):3567--3572.

\bibitem[{Jiang et~al.(2023)Jiang, Sablayrolles, Mensch, Bamford, Chaplot, de~las Casas, Bressand, Lengyel, Lample, Saulnier et~al.}]{jiang2023mistral}
Albert~Q. Jiang, Alexandre Sablayrolles, Arthur Mensch, Chris Bamford, Devendra~Singh Chaplot, Diego de~las Casas, Florian Bressand, Gianna Lengyel, Guillaume Lample, Lucile Saulnier, and 1 others. 2023.
\newblock \href {https://arxiv.org/pdf/2310.06825} {Mistral 7b}.
\newblock \emph{arXiv preprint arXiv:2310.06825}.

\bibitem[{Jin et~al.(2020)Jin, Barzilay, and Jaakkola}]{jin2020multi}
Wengong Jin, Regina Barzilay, and Tommi Jaakkola. 2020.
\newblock \href {http://proceedings.mlr.press/v119/jin20b/jin20b.pdf} {Multi-objective molecule generation using interpretable substructures}.
\newblock In \emph{International Conference on Machine Learning}, pages 4849--4859. PMLR.

\bibitem[{Kim et~al.(2019)Kim, Chen, Cheng, Gindulyte, He, He, Li, Shoemaker, Thiessen, Yu et~al.}]{kim2019pubchem}
Sunghwan Kim, Jie Chen, Tiejun Cheng, Asta Gindulyte, Jia He, Siqian He, Qingliang Li, Benjamin~A Shoemaker, Paul~A Thiessen, Bo~Yu, and 1 others. 2019.
\newblock \href {https://academic.oup.com/nar/article-pdf/47/D1/D1102/27437306/gky1033.pdf} {Pubchem 2019 update: improved access to chemical data}.
\newblock \emph{Nucleic acids research}, 47(D1):D1102--D1109.

\bibitem[{Kim et~al.(2025)Kim, Chen, Cheng, Gindulyte, He, He, Li, Shoemaker, Thiessen, Yu et~al.}]{kim2025pubchem}
Sunghwan Kim, Jie Chen, Tiejun Cheng, Asta Gindulyte, Jia He, Siqian He, Qingliang Li, Benjamin~A Shoemaker, Paul~A Thiessen, Bo~Yu, and 1 others. 2025.
\newblock \href {https://academic.oup.com/nar/article-pdf/53/D1/D1516/60743708/gkae1059.pdf} {Pubchem 2025 update}.
\newblock \emph{Nucleic Acids Research}, 53(D1):D1516--D1525.

\bibitem[{Kim et~al.(2016)Kim, Thiessen, Bolton, Chen, Fu, Gindulyte, Han, He, He, Shoemaker et~al.}]{kim2016pubchem}
Sunghwan Kim, Paul~A Thiessen, Evan~E Bolton, Jie Chen, Gang Fu, Asta Gindulyte, Lianyi Han, Jane He, Siqian He, Benjamin~A Shoemaker, and 1 others. 2016.
\newblock \href {https://academic.oup.com/nar/article-pdf/44/D1/D1202/9484096/gkv951.pdf} {Pubchem substance and compound databases}.
\newblock \emph{Nucleic acids research}, 44(D1):D1202--D1213.

\bibitem[{Krenn et~al.(2020)Krenn, H{\"a}se, Nigam, Friederich, and Aspuru-Guzik}]{krenn2020self}
Mario Krenn, Florian H{\"a}se, AkshatKumar Nigam, Pascal Friederich, and Alan Aspuru-Guzik. 2020.
\newblock \href {https://iopscience.iop.org/article/10.1088/2632-2153/aba947/pdf} {Self-referencing embedded strings (selfies): A 100\% robust molecular string representation}.
\newblock \emph{Machine Learning: Science and Technology}, 1(4):045024.

\bibitem[{Landrum et~al.(2013)}]{landrum2013rdkit}
Greg Landrum and 1 others. 2013.
\newblock \href {http://www.rdkit.org/RDKit_Overview.pdf} {Rdkit: A software suite for cheminformatics, computational chemistry, and predictive modeling}.
\newblock \emph{Greg Landrum}, 8(31.10):5281.

\bibitem[{Le and Chawla(2024)}]{le2024utilizing}
Khiem Le and Nitesh~V Chawla. 2024.
\newblock \href {https://arxiv.org/pdf/2410.13147} {Utilizing large language models in an iterative paradigm with domain feedback for molecule optimization}.
\newblock \emph{arXiv preprint arXiv:2410.13147}.

\bibitem[{Le et~al.(2024)Le, Guo, Dong, Huang, Nan, Iyer, Zhang, Wiest, Wang, and Chawla}]{le2024molx}
Khiem Le, Zhichun Guo, Kaiwen Dong, Xiaobao Huang, Bozhao Nan, Roshni Iyer, Xiangliang Zhang, Olaf Wiest, Wei Wang, and Nitesh~V Chawla. 2024.
\newblock \href {https://arxiv.org/pdf/2406.06777} {Molx: Enhancing large language models for molecular learning with a multi-modal extension}.
\newblock \emph{arXiv preprint arXiv:2406.06777}.

\bibitem[{Lee et~al.(2025)Lee, Song, Jeong, Ko, Hormazabal, Han, Bae, Lim, and Kim}]{lee2025mol}
Chanhui Lee, Yuheon Song, YongJun Jeong, Hanbum Ko, Rodrigo Hormazabal, Sehui Han, Kyunghoon Bae, Sungbin Lim, and Sungwoong Kim. 2025.
\newblock \href {https://arxiv.org/pdf/2502.02810} {Mol-llm: Generalist molecular llm with improved graph utilization}.
\newblock \emph{arXiv preprint arXiv:2502.02810}.

\bibitem[{Levenshtein(1966)}]{levenshtein1966binary}
Vladimir~I Levenshtein. 1966.
\newblock \href {https://nymity.ch/sybilhunting/pdf/Levenshtein1966a.pdf} {Binary codes capable of correcting deletions, insertions, and reversals}.
\newblock \emph{Soviet physics doklady}, 10(8):707--710.

\bibitem[{Li et~al.(2024{\natexlab{a}})Li, Li, Liu, Zhou, and Li}]{li2024tomg}
Jiatong Li, Junxian Li, Yunqing Liu, Dongzhan Zhou, and Qing Li. 2024{\natexlab{a}}.
\newblock \href {https://arxiv.org/pdf/2412.14642} {Tomg-bench: Evaluating llms on text-based open molecule generation}.
\newblock \emph{arXiv preprint arXiv:2412.14642}.

\bibitem[{Li et~al.(2024{\natexlab{b}})Li, Liu, Ding, Fan, Li, and Li}]{li2024large}
Jiatong Li, Wei Liu, Zhihao Ding, Wenqi Fan, Yuqiang Li, and Qing Li. 2024{\natexlab{b}}.
\newblock \href {https://ieeexplore.ieee.org/iel8/69/4358933/10948482.pdf} {Large language models are in-context molecule learners}.
\newblock \emph{arXiv preprint arXiv:2403.04197}.

\bibitem[{Li et~al.(2024{\natexlab{c}})Li, Liu, Fan, Wei, Liu, Tang, and Li}]{li2024empowering}
Jiatong Li, Yunqing Liu, Wenqi Fan, Xiao-Yong Wei, Hui Liu, Jiliang Tang, and Qing Li. 2024{\natexlab{c}}.
\newblock \href {https://ieeexplore.ieee.org/iel7/69/4358933/10516270.pdf} {Empowering molecule discovery for molecule-caption translation with large language models: A chatgpt perspective}.
\newblock \emph{IEEE transactions on knowledge and data engineering}.

\bibitem[{Li et~al.(2024{\natexlab{d}})Li, Liu, Liu, Le, Zhang, Fan, Zhou, Li, and Li}]{li2024molreflect}
Jiatong Li, Yunqing Liu, Wei Liu, Jingdi Le, Di~Zhang, Wenqi Fan, Dongzhan Zhou, Yuqiang Li, and Qing Li. 2024{\natexlab{d}}.
\newblock \href {https://arxiv.org/pdf/2411.14721} {Molreflect: Towards in-context fine-grained alignments between molecules and texts}.
\newblock \emph{arXiv preprint arXiv:2411.14721}.

\bibitem[{Liao et~al.(2024)Liao, Yu, Mei, and Wei}]{liao2024words}
Chang Liao, Yemin Yu, Yu~Mei, and Ying Wei. 2024.
\newblock \href {https://arxiv.org/pdf/2402.01439} {From words to molecules: A survey of large language models in chemistry}.
\newblock \emph{arXiv preprint arXiv:2402.01439}.

\bibitem[{Lin et~al.(2024)Lin, Xia, Huang, Liu, Zhang, and Gao}]{lin2024versatile}
Xiaohan Lin, Yijie Xia, Yupeng Huang, Shuo Liu, Jun Zhang, and Yi~Qin Gao. 2024.
\newblock \href {https://chemrxiv.org/engage/api-gateway/chemrxiv/assets/orp/resource/item/66702f4401103d79c565deb4/original/versatile-molecular-editing-via-multimodal-and-group-optimized-generative-learning.pdf} {Versatile molecular editing via multimodal and group-optimized generative learning}.

\bibitem[{Lin et~al.(2025)Lin, Chen, Wang, Zeng, and Yu}]{lin2025prop}
Xuan Lin, Long Chen, Yile Wang, Xiangxiang Zeng, and Philip~S. Yu. 2025.
\newblock \href {https://arxiv.org/pdf/2412.18084} {Property enhanced instruction tuning for multi-task molecule generation with large language models}.
\newblock \emph{arXiv preprint arXiv:2412.18084}.

\bibitem[{Liu et~al.(2025{\natexlab{a}})Liu, Li, Zhang, Wang, He, Hong, Liu, Zhang, Song, Zhu et~al.}]{liu2025advances}
Bang Liu, Xinfeng Li, Jiayi Zhang, Jinlin Wang, Tanjin He, Sirui Hong, Hongzhang Liu, Shaokun Zhang, Kaitao Song, Kunlun Zhu, and 1 others. 2025{\natexlab{a}}.
\newblock \href {https://arxiv.org/pdf/2504.01990} {Advances and challenges in foundation agents: From brain-inspired intelligence to evolutionary, collaborative, and safe systems}.
\newblock \emph{arXiv preprint arXiv:2504.01990}.

\bibitem[{Liu et~al.(2024{\natexlab{a}})Liu, Sun, Matusik, Jiang, and Chen}]{liu2024multimodal}
Gang Liu, Michael Sun, Wojciech Matusik, Meng Jiang, and Jie Chen. 2024{\natexlab{a}}.
\newblock \href {https://arxiv.org/pdf/2410.04223} {Multimodal large language models for inverse molecular design with retrosynthetic planning}.
\newblock \emph{arXiv preprint arXiv:2410.04223}.

\bibitem[{Liu et~al.(2024{\natexlab{b}})Liu, Tao, and Ren}]{liu2024scientific}
Pengfei Liu, Jun Tao, and Zhixiang Ren. 2024{\natexlab{b}}.
\newblock \href {https://www.researchgate.net/profile/Pengfei-Liu-75/publication/380879218_Scientific_Language_Modeling_A_Quantitative_Review_of_Large_Language_Models_in_Molecular_Science/links/665295f0479366623a130db3/Scientific-Language-Modeling-A-Quantitative-Review-of-Large-Language-Models-in-Molecular-Science.pdf} {Scientific language modeling: A quantitative review of large language models in molecular science}.
\newblock \emph{arXiv preprint arXiv:2402.04119}, page~3.

\bibitem[{Liu et~al.(2025{\natexlab{b}})Liu, Tao, and Ren}]{liu2025quantitative}
Pengfei Liu, Jun Tao, and Zhixiang Ren. 2025{\natexlab{b}}.
\newblock \href {https://arxiv.org/pdf/2402.04119} {A quantitative analysis of knowledge-learning preferences in large language models in molecular science}.
\newblock \emph{Nature Machine Intelligence}, pages 1--13.

\bibitem[{Liu et~al.(2023)Liu, Nie, Wang, Lu, Qiao, Liu, Tang, Xiao, and Anandkumar}]{liu2023multi}
Shengchao Liu, Weili Nie, Chengpeng Wang, Jiarui Lu, Zhuoran Qiao, Ling Liu, Jian Tang, Chaowei Xiao, and Animashree Anandkumar. 2023.
\newblock \href {https://arxiv.org/pdf/2212.10789} {Multi-modal molecule structure--text model for text-based retrieval and editing}.
\newblock \emph{Nature Machine Intelligence}, 5(12):1447--1457.

\bibitem[{Liu et~al.(2024{\natexlab{c}})Liu, Wang, Yang, Wang, Liu, Guo, and Xiao}]{liu2024conversational}
Shengchao Liu, Jiongxiao Wang, Yijin Yang, Chengpeng Wang, Ling Liu, Hongyu Guo, and Chaowei Xiao. 2024{\natexlab{c}}.
\newblock \href {https://openreview.net/pdf?id=yRrPfKyJQ2} {Conversational drug editing using retrieval and domain feedback}.
\newblock In \emph{The twelfth international conference on learning representations}.

\bibitem[{Liu et~al.(2024{\natexlab{d}})Liu, Guo, Li, Liu, Huang, Ke, and Lv}]{liu2024drugllm}
Xianggen Liu, Yan Guo, Haoran Li, Jin Liu, Shudong Huang, Bowen Ke, and Jiancheng Lv. 2024{\natexlab{d}}.
\newblock \href {https://arxiv.org/pdf/2405.06690} {Drugllm: Open large language model for few-shot molecule generation}.
\newblock \emph{arXiv preprint arXiv:2405.06690}.

\bibitem[{Liu et~al.(2025{\natexlab{c}})Liu, Jiang, Chen, Yang, Chen, Foster, and Stevens}]{liu2025drugimprovergpt}
Xuefeng Liu, Songhao Jiang, Siyu Chen, Zhuoran Yang, Yuxin Chen, Ian Foster, and Rick Stevens. 2025{\natexlab{c}}.
\newblock \href {https://arxiv.org/pdf/2502.07237} {Drugimprovergpt: A large language model for drug optimization with fine-tuning via structured policy optimization}.
\newblock \emph{arXiv preprint arXiv:2502.07237}.

\bibitem[{Liu et~al.(2025{\natexlab{d}})Liu, Jiang, Li, and Stevens}]{liu2025controllablegpt}
Xuefeng Liu, Songhao Jiang, Bo~Li, and Rick Stevens. 2025{\natexlab{d}}.
\newblock \href {https://arxiv.org/pdf/2502.10631?} {Controllablegpt: A ground-up designed controllable gpt for molecule optimization}.
\newblock \emph{arXiv preprint arXiv:2502.10631}.

\bibitem[{Liu et~al.(2025{\natexlab{e}})Liu, Jiang, and Stevens}]{liu2025scaffoldgpt}
Xuefeng Liu, Songhao Jiang, and Rick Stevens. 2025{\natexlab{e}}.
\newblock \href {https://arxiv.org/pdf/2502.06891} {Scaffoldgpt: A scaffold-based large language model for drug improvement}.
\newblock \emph{arXiv preprint arXiv:2502.06891}.

\bibitem[{Lu et~al.(2023)Lu, Wei, Wang, Zhang, and Liu}]{lu2023graphgpt}
Hao Lu, Zhiqiang Wei, Xuze Wang, Kun Zhang, and Hao Liu. 2023.
\newblock \href {https://www.mdpi.com/1422-0067/24/23/16761} {Graphgpt: A graph enhanced generative pretrained transformer for conditioned molecular generation}.
\newblock \emph{International Journal of Molecular Sciences}, 24(23):16761.

\bibitem[{Lu et~al.(2024)Lu, Song, Zhao, Du, Cao, Jia, and Duan}]{lu2024generative}
Jieyu Lu, Zhangde Song, Qiyuan Zhao, Yuanqi Du, Yirui Cao, Haojun Jia, and Chenru Duan. 2024.
\newblock \href {https://doi.org/10.48550/arXiv.2410.18136} {Generative design of functional metal complexes utilizing the internal knowledge of large language models}.
\newblock \emph{arXiv preprint arXiv:2410.18136}.

\bibitem[{Ma et~al.(2025)Ma, He, Snell, Griggs, Min, and Zaharia}]{ma2025reasoning}
Wenjie Ma, Jingxuan He, Charlie Snell, Tyler Griggs, Sewon Min, and Matei Zaharia. 2025.
\newblock \href {https://arxiv.org/pdf/2504.09858?} {Reasoning models can be effective without thinking}.
\newblock \emph{arXiv preprint arXiv:2504.09858}.

\bibitem[{Mirza et~al.(2024)Mirza, Alampara, Kunchapu, Emoekabu, Krishnan, Wilhelmi, Okereke, Eberhardt, Elahi, Greiner et~al.}]{mirza2404large}
A~Mirza, N~Alampara, S~Kunchapu, B~Emoekabu, A~Krishnan, M~Wilhelmi, M~Okereke, J~Eberhardt, AM~Elahi, M~Greiner, and 1 others. 2024.
\newblock \href {https://arxiv.org/pdf/2404.01475} {Are large language models superhuman chemists?}
\newblock \emph{arXiv preprint arXiv:2404.01475}.

\bibitem[{Morgan(1965)}]{morgan1965generation}
HL~Morgan. 1965.
\newblock \href {https://pubs.acs.org/doi/pdf/10.1021/c160017a018} {The generation of a unique machine description for chemical structures—a technique developed at chemical abstracts service}.
\newblock \emph{Journal of Chemical Documentation}, 5(2):107--113.

\bibitem[{Nakamura et~al.(2025)Nakamura, Yasuo, and Sekijima}]{nakamura2025molecular}
Shogo Nakamura, Nobuaki Yasuo, and Masakazu Sekijima. 2025.
\newblock \href {https://www.nature.com/articles/s42004-025-01437-x.pdf} {Molecular optimization using a conditional transformer for reaction-aware compound exploration with reinforcement learning}.
\newblock \emph{Communications Chemistry}, 8(1):40.

\bibitem[{Papineni et~al.(2002)Papineni, Roukos, Ward, and Zhu}]{papineni2002bleu}
Kishore Papineni, Salim Roukos, Todd Ward, and Wei-Jing Zhu. 2002.
\newblock \href {https://aclanthology.org/P02-1040.Pdf} {Bleu: a method for automatic evaluation of machine translation}.
\newblock In \emph{Proceedings of the 40th annual meeting of the Association for Computational Linguistics}, pages 311--318.

\bibitem[{Pareto(1919)}]{pareto1919manuale}
Vilfredo Pareto. 1919.
\newblock \href {https://books.google.com/books?hl=en&lr=&id=osYOAQAAIAAJ&oi=fnd&pg=PA1&dq=Manuale+di+Economia+Politica&ots=SVHSUPMV65&sig=1xrKooXHRm027LKfPtIDte0DRDI} {\emph{Manuale di economia politica con una introduzione alla scienza sociale}}, volume~13.
\newblock Societ{\`a} editrice libraria.

\bibitem[{Park et~al.(2025)Park, Ahn, Choi, and Kim}]{park2025mol}
Jinyeong Park, Jaegyoon Ahn, Jonghwan Choi, and Jibum Kim. 2025.
\newblock \href {https://pubs.acs.org/doi/pdf/10.1021/acs.jcim.4c01669} {Mol-air: Molecular reinforcement learning with adaptive intrinsic rewards for goal-directed molecular generation}.
\newblock \emph{Journal of Chemical Information and Modeling}, 65(5):2283--2296.

\bibitem[{Pinheiro et~al.(2020)Pinheiro, Mucelini, Soares, Prati, Da~Silva, and Quiles}]{pinheiro2020machine}
Gabriel~A Pinheiro, Johnatan Mucelini, Marinalva~D Soares, Ronaldo~C Prati, Juarez~LF Da~Silva, and Marcos~G Quiles. 2020.
\newblock \href {https://pubs.acs.org/doi/pdf/10.1021/acs.jpca.0c05969} {Machine learning prediction of nine molecular properties based on the smiles representation of the qm9 quantum-chemistry dataset}.
\newblock \emph{The Journal of Physical Chemistry A}, 124(47):9854--9866.

\bibitem[{Pirnay et~al.(2025)Pirnay, Rittig, Wolf, Grohe, Burger, Mitsos, and Grimm}]{pirnay2025graphxform}
Jonathan Pirnay, Jan~G Rittig, Alexander~B Wolf, Martin Grohe, Jakob Burger, Alexander Mitsos, and Dominik~G Grimm. 2025.
\newblock \href {https://pubs.rsc.org/en/content/articlepdf/2025/dd/d4dd00339j} {Graphxform: graph transformer for computer-aided molecular design}.
\newblock \emph{Digital Discovery}, 4(4):1052--1065.

\bibitem[{Polykovskiy et~al.(2020)Polykovskiy, Zhebrak, Sanchez-Lengeling, Golovanov, Tatanov, Belyaev, Kurbanov, Artamonov, Aladinskiy, Veselov et~al.}]{polykovskiy2020molecular}
Daniil Polykovskiy, Alexander Zhebrak, Benjamin Sanchez-Lengeling, Sergey Golovanov, Oktai Tatanov, Stanislav Belyaev, Rauf Kurbanov, Aleksey Artamonov, Vladimir Aladinskiy, Mark Veselov, and 1 others. 2020.
\newblock \href {https://www.frontiersin.org/articles/10.3389/fphar.2020.565644/pdf} {Molecular sets (moses): a benchmarking platform for molecular generation models}.
\newblock \emph{Frontiers in pharmacology}, 11:565644.

\bibitem[{Preuer et~al.(2018)Preuer, Renz, Unterthiner et~al.}]{preuer2018frechet}
Kristina Preuer, Philipp Renz, Thomas Unterthiner, and 1 others. 2018.
\newblock \href {https://pubs.acs.org/doi/pdf/10.1021/acs.jcim.8b00234} {Fr{\'e}chet chemnet distance: A metric for generative models for molecules in drug discovery}.
\newblock \emph{Journal of chemical information and modeling}, 58(9):1736--1741.

\bibitem[{Rajpurkar et~al.(2016)Rajpurkar, Zhang, Lopyrev, and Liang}]{rajpurkar2016squad}
Pranav Rajpurkar, Jian Zhang, Konstantin Lopyrev, and Percy Liang. 2016.
\newblock \href {https://arxiv.org/pdf/1606.05250} {Squad: 100,000+ questions for machine comprehension of text}.
\newblock \emph{arXiv preprint arXiv:1606.05250}.

\bibitem[{Ramos et~al.(2025)Ramos, Collison, and White}]{ramos2025review}
Mayk~Caldas Ramos, Christopher~J. Collison, and Andrew~D. White. 2025.
\newblock \href {https://pubs.rsc.org/en/content/articlepdf/2024/sc/d4sc03921a} {A review of large language models and autonomous agents in chemistry}.
\newblock \emph{Chemical Science}.

\bibitem[{Ran et~al.(2025)Ran, Wang, and Allmendinger}]{ran2025multi}
Nian Ran, Yue Wang, and Richard Allmendinger. 2025.
\newblock \href {https://doi.org/10.48550/arXiv.2502.12845} {{MOLLM:} multi-objective large language model for molecular design - optimizing with experts}.
\newblock \emph{arXiv preprint arXiv:2502.12845}.

\bibitem[{Ross et~al.(2022)Ross, Belgodere, Chenthamarakshan, Padhi, Mroueh, and Das}]{ross2022large}
Jerret Ross, Brian Belgodere, Vijil Chenthamarakshan, Inkit Padhi, Youssef Mroueh, and Payel Das. 2022.
\newblock \href {https://arxiv.org/pdf/2106.09553} {Large-scale chemical language representations capture molecular structure and properties}.
\newblock \emph{Nature Machine Intelligence}, 4(12):1256--1264.

\bibitem[{Ross et~al.(2024)Ross, Hoffman, Belgodere, Chenthamarakshan, Mroueh, and Das}]{ross2024learning}
Jerret Ross, Samuel Hoffman, Brian Belgodere, Vijil Chenthamarakshan, Youssef Mroueh, and Payel Das. 2024.
\newblock \href {https://openreview.net/pdf?id=FhxDKumovH} {Learning to optimize molecules with a chemical language model}.
\newblock In \emph{Annual Conference on Neural Information Processing Systems}.

\bibitem[{Srinivas and Runkana(2024)}]{srinivas2024cross}
Sakhinana~Sagar Srinivas and Venkataramana Runkana. 2024.
\newblock \href {https://arxiv.org/pdf/2408.11866} {Crossing new frontiers: Knowledge-augmented large language model prompting for zero-shot text-based de novo molecule design}.
\newblock \emph{arXiv preprint arXiv:2408.11866}.

\bibitem[{Sun et~al.(2025)Sun, Bagni, Cavanagh, Wang, Sawyer, Gritsevskiy, Zhang, and Head-Gordon}]{sun2025synllama}
Kunyang Sun, Dorian Bagni, Joseph~M Cavanagh, Yingze Wang, Jacob~M Sawyer, Andrew Gritsevskiy, Oufan Zhang, and Teresa Head-Gordon. 2025.
\newblock \href {https://arxiv.org/pdf/2503.12602} {Synllama: Generating synthesizable molecules and their analogs with large language models}.
\newblock \emph{arXiv preprint arXiv:2503.12602}.

\bibitem[{Sutton et~al.(1998)Sutton, Barto et~al.}]{sutton1998reinforcement}
Richard~S Sutton, Andrew~G Barto, and 1 others. 1998.
\newblock \href {https://www.academia.edu/download/38529120/9780262257053_index.pdf} {\emph{Reinforcement learning: An introduction}}, volume~1.
\newblock MIT press Cambridge.

\bibitem[{Tang et~al.(2024)Tang, Dai, Knight, Wu, Li, Li, and Gerstein}]{tang2024survey}
Xiangru Tang, Howard Dai, Elizabeth Knight, Fang Wu, Yunyang Li, Tianxiao Li, and Mark Gerstein. 2024.
\newblock \href {https://academic.oup.com/bib/article-pdf/25/4/bbae338/58544272/bbae338.pdf} {A survey of generative ai for de novo drug design: new frontiers in molecule and protein generation}.
\newblock \emph{Briefings in Bioinformatics}, 25(4):bbae338.

\bibitem[{Touvron et~al.(2023)Touvron, Lavril, Izacard, Martinet, Lachaux, Lacroix, Rozi{\`e}re, Goyal, Hambro, Azhar et~al.}]{touvron2023llama}
Hugo Touvron, Thibaut Lavril, Gautier Izacard, Xavier Martinet, Marie-Anne Lachaux, Timoth{\'e}e Lacroix, Baptiste Rozi{\`e}re, Naman Goyal, Eric Hambro, Faisal Azhar, and 1 others. 2023.
\newblock \href {https://glossary.midtown.ai/assets/l/llama_model_paper.pdf} {Llama: Open and efficient foundation language models}.
\newblock \emph{arXiv preprint arXiv:2302.13971}.

\bibitem[{Wang et~al.(2025)Wang, Skreta, Ser, Gao, Kong, Strieth-Kalthoff, Duan, Zhuang, Yu, Zhu, Du, Aspuru-Guzik, Neklyudov, and Zhang}]{wang2025efficient}
Haorui Wang, Marta Skreta, Cher~Tian Ser, Wenhao Gao, Lingkai Kong, Felix Strieth-Kalthoff, Chenru Duan, Yuchen Zhuang, Yue Yu, Yanqiao Zhu, Yuanqi Du, Alan Aspuru-Guzik, Kirill Neklyudov, and Chao Zhang. 2025.
\newblock \href {https://openreview.net/forum?id=awWiNvQwf3} {Efficient evolutionary search over chemical space with large language models}.
\newblock In \emph{The Thirteenth International Conference on Learning Representations}.

\bibitem[{Wang et~al.(2023)Wang, Zhao, Sciabola, and Wang}]{wang2023cmol}
Y.~Wang, H.~Zhao, S.~Sciabola, and W.~Wang. 2023.
\newblock \href {https://www.mdpi.com/1420-3049/28/11/4430/pdf} {cmolgpt: A conditional generative pre-trained transformer for target-specific de novo molecular generation}.
\newblock \emph{Molecules}, 28(11):4430.

\bibitem[{Wei et~al.(2022{\natexlab{a}})Wei, Tay, Bommasani, Raffel, Zoph, Borgeaud, Yogatama, Bosma, Zhou, Metzler et~al.}]{wei2022emergent}
Jason Wei, Yi~Tay, Rishi Bommasani, Colin Raffel, Barret Zoph, Sebastian Borgeaud, Dani Yogatama, Maarten Bosma, Denny Zhou, Donald Metzler, and 1 others. 2022{\natexlab{a}}.
\newblock \href {https://arxiv.org/pdf/2206.07682} {Emergent abilities of large language models}.
\newblock \emph{TMLR}.

\bibitem[{Wei et~al.(2022{\natexlab{b}})Wei, Wang, Schuurmans, Bosma, Xia, Chi, Le, Zhou et~al.}]{wei2022chain}
Jason Wei, Xuezhi Wang, Dale Schuurmans, Maarten Bosma, Fei Xia, Ed~Chi, Quoc~V Le, Denny Zhou, and 1 others. 2022{\natexlab{b}}.
\newblock \href {https://proceedings.neurips.cc/paper_files/paper/2022/file/9d5609613524ecf4f15af0f7b31abca4-Paper-Conference.pdf} {Chain-of-thought prompting elicits reasoning in large language models}.
\newblock \emph{NeurIPS}, 35:24824--24837.

\bibitem[{Weininger(1988)}]{weininger1988smiles}
David Weininger. 1988.
\newblock \href {https://elearning.uniroma1.it/pluginfile.php/728447/mod_folder/content/0/PDF.ArticoliSupporto/1988.SMILES,%20a%20chemical%20language%20and%20information%20system.%201.%20Introduction%20to%20methodology%20and%20encoding%20rules.pdf} {Smiles, a chemical language and information system. 1. introduction to methodology and encoding rules}.
\newblock \emph{Journal of chemical information and computer sciences}, 28(1):31--36.

\bibitem[{Wu et~al.(2018)Wu, Ramsundar, Feinberg, Gomes, Geniesse, Pappu, Leswing, and Pande}]{wu2018moleculenet}
Zhenqin Wu, Bharath Ramsundar, Evan~N Feinberg, Joseph Gomes, Caleb Geniesse, Aneesh~S Pappu, Karl Leswing, and Vijay Pande. 2018.
\newblock \href {https://pubs.rsc.org/de-at/content/articlepdf/2018/sc/c7sc02664a} {Moleculenet: a benchmark for molecular machine learning}.
\newblock \emph{Chemical science}, 9(2):513--530.

\bibitem[{Wu et~al.(2024)Wu, Zhang, Wang, Fu, Zhao, Wang, Du, Jiang, Deng, Cao et~al.}]{wu2024leveraging}
Zhenxing Wu, Odin Zhang, Xiaorui Wang, Li~Fu, Huifeng Zhao, Jike Wang, Hongyan Du, Dejun Jiang, Yafeng Deng, Dongsheng Cao, and 1 others. 2024.
\newblock \href {https://www.nature.com/articles/s42256-024-00916-5} {Leveraging language model for advanced multiproperty molecular optimization via prompt engineering}.
\newblock \emph{Nature Machine Intelligence}, pages 1--11.

\bibitem[{Xia et~al.(2025)Xia, Jin, Xie, He, Cao, Luo, Liu, Wang, Liu, Chen et~al.}]{xia2025naturelm}
Yingce Xia, Peiran Jin, Shufang Xie, Liang He, Chuan Cao, Renqian Luo, Guoqing Liu, Yue Wang, Zequn Liu, Yuan-Jyue Chen, and 1 others. 2025.
\newblock \href {https://arxiv.org/pdf/2502.07527} {Naturelm: Deciphering the language of nature for scientific discovery}.
\newblock \emph{arXiv preprint arXiv:2502.07527}.

\bibitem[{Xiao et~al.(2025)Xiao, Cai, Wang, and Zhou}]{xiao2025m}
Meng Xiao, Xunxin Cai, Chengrui Wang, and Yuanchun Zhou. 2025.
\newblock \href {https://arxiv.org/pdf/2504.19565} {m-kailin: Knowledge-driven agentic scientific corpus distillation framework for biomedical large language models training}.
\newblock \emph{arXiv preprint arXiv:2504.19565}.

\bibitem[{Xu et~al.(2024)Xu, Sharaf, Chen, Tan, Shen, Van~Durme, Murray, and Kim}]{xu2024contrastive}
Haoran Xu, Amr Sharaf, Yunmo Chen, Weiting Tan, Lingfeng Shen, Benjamin Van~Durme, Kenton Murray, and Young~Jin Kim. 2024.
\newblock \href {https://arxiv.org/pdf/2401.08417} {Contrastive preference optimization: Pushing the boundaries of llm performance in machine translation}.
\newblock \emph{arXiv preprint arXiv:2401.08417}.

\bibitem[{Yang et~al.(2023)Yang, Jin, Tang, Han, Feng, Jiang, Yin, and Hu}]{yang2023harnessing}
Jingfeng Yang, Hongye Jin, Ruixiang Tang, Xiaotian Han, Qizhang Feng, Haoming Jiang, Bing Yin, and Xia Hu. 2023.
\newblock \href {https://dl.acm.org/doi/pdf/10.1145/3649506} {Harnessing the power of llms in practice: A survey on chatgpt and beyond}.
\newblock \emph{arXiv preprint arXiv:2304.13712}.

\bibitem[{Yang et~al.(2024)Yang, Wu, Zeng, Li, Bao, and Yan}]{yang2024molecule}
Nianzu Yang, Huaijin Wu, Kaipeng Zeng, Yang Li, Siyuan Bao, and Junchi Yan. 2024.
\newblock \href {https://www.sciencedirect.com/science/article/pii/S2667325824005259} {Molecule generation for drug design: a graph learning perspective}.
\newblock \emph{Fundamental Research}.

\bibitem[{Yao et~al.(2024)Yao, Wang, Zhang, Qin, Zhang, Chu, Yang, Zhu, and Mei}]{yao2024exploring}
Yang Yao, Xin Wang, Zeyang Zhang, Yijian Qin, Ziwei Zhang, Xu~Chu, Yuekui Yang, Wenwu Zhu, and Hong Mei. 2024.
\newblock \href {https://arxiv.org/pdf/2403.14358} {Exploring the potential of large language models in graph generation}.
\newblock \emph{arXiv preprint arXiv:2403.14358}.

\bibitem[{Ye et~al.(2025)Ye, Cai, Lai, Wang, Huang, Wang, Liu, and Zeng}]{ye2025drugassist}
Geyan Ye, Xibao Cai, Houtim Lai, Xing Wang, Junhong Huang, Longyue Wang, Wei Liu, and Xiangxiang Zeng. 2025.
\newblock \href {https://academic.oup.com/bib/article-pdf/26/1/bbae693/61326352/bbae693.pdf} {Drugassist: A large language model for molecule optimization}.
\newblock \emph{Briefings in Bioinformatics}, 26(1):bbae693.

\bibitem[{Yu et~al.(2024{\natexlab{a}})Yu, Baker, Chen, Ning, and Sun}]{yu2024llasmol}
Botao Yu, Frazier~N Baker, Ziqi Chen, Xia Ning, and Huan Sun. 2024{\natexlab{a}}.
\newblock \href {https://arxiv.org/pdf/2402.09391} {Llasmol: Advancing large language models for chemistry with a large-scale, comprehensive, high-quality instruction tuning dataset}.
\newblock \emph{arXiv preprint arXiv:2402.09391}.

\bibitem[{Yu et~al.(2024{\natexlab{b}})Yu, Baker, Chen, Ning, and Sun}]{yu2024advancing}
Botao Yu, Frazier~N. Baker, Ziqi Chen, Xia Ning, and Huan Sun. 2024{\natexlab{b}}.
\newblock \href {https://doi.org/10.48550/arXiv.2402.09391} {Llasmol: Advancing large language models for chemistry with a large-scale, comprehensive, high-quality instruction tuning dataset}.
\newblock \emph{arXiv preprint arXiv:2402.09391}.

\bibitem[{Yu et~al.(2025)Yu, Zheng, Koh, Pan, Wang, and Wang}]{yu2025collaborative}
Jiajun Yu, Yizhen Zheng, Huan~Yee Koh, Shirui Pan, Tianyue Wang, and Haishuai Wang. 2025.
\newblock \href {https://arxiv.org/pdf/2503.03503?} {Collaborative expert llms guided multi-objective molecular optimization}.
\newblock \emph{arXiv preprint arXiv:2503.03503}.

\bibitem[{Yuan and F{\"a}rber(2025)}]{yuan2025hallucinations}
Shuzhou Yuan and Michael F{\"a}rber. 2025.
\newblock \href {https://arxiv.org/pdf/2501.13824} {Hallucinations can improve large language models in drug discovery}.
\newblock \emph{arXiv preprint arXiv:2501.13824}.

\bibitem[{Zeng et~al.(2022)Zeng, Wang, Luo, Kang, Tang, Lightstone, Fang, Cornell, Nussinov, and Cheng}]{zeng2022deep}
Xiangxiang Zeng, Fei Wang, Yuan Luo, Seung-gu Kang, Jian Tang, Felice~C Lightstone, Evandro~F Fang, Wendy Cornell, Ruth Nussinov, and Feixiong Cheng. 2022.
\newblock \href {https://www.cell.com/cell-reports-medicine/pdf/S2666-3791%2822%2900349-4.pdf} {Deep generative molecular design reshapes drug discovery}.
\newblock \emph{Cell Reports Medicine}, 3(12).

\bibitem[{Zhang et~al.(2024{\natexlab{a}})Zhang, Liu, Tan, Chen, Yan, Yan, Li, Huang, Yue, Ouyang et~al.}]{zhang2024chemllm}
Di~Zhang, Wei Liu, Qian Tan, Jingdan Chen, Hang Yan, Yuliang Yan, Jiatong Li, Weiran Huang, Xiangyu Yue, Wanli Ouyang, and 1 others. 2024{\natexlab{a}}.
\newblock \href {https://arxiv.org/pdf/2402.06852} {Chemllm: A chemical large language model}.
\newblock \emph{arXiv preprint arXiv:2402.06852}.

\bibitem[{Zhang et~al.(2024{\natexlab{b}})Zhang, Bian, Chen, and Yao}]{zhang2024unimot}
Juzheng Zhang, Yatao Bian, Yongqiang Chen, and Quanming Yao. 2024{\natexlab{b}}.
\newblock \href {https://arxiv.org/pdf/2408.00863} {Unimot: Unified molecule-text language model with discrete token representation.}
\newblock \emph{arXiv preprint arXiv:2408.00863}.

\bibitem[{Zhang et~al.(2024{\natexlab{c}})Zhang, Lin, Zhang, Zhao, Huang, Hsieh, Pan, and Hou}]{zhang2024deep}
Odin Zhang, Haitao Lin, Hui Zhang, Huifeng Zhao, Yufei Huang, Chang-Yu Hsieh, Peichen Pan, and Tingjun Hou. 2024{\natexlab{c}}.
\newblock \href {https://arxiv.org/pdf/2404.19230} {Deep lead optimization: Leveraging generative ai for structural modification}.
\newblock \emph{Journal of the American Chemical Society}, 146(46):31357--31370.

\bibitem[{Zhang et~al.(2025)Zhang, Ding, Lv, Wang, Yin, Zhang, Yu, Wang, Li, Xiang et~al.}]{zhang2025scientific}
Qiang Zhang, Keyan Ding, Tianwen Lv, Xinda Wang, Qingyu Yin, Yiwen Zhang, Jing Yu, Yuhao Wang, Xiaotong Li, Zhuoyi Xiang, and 1 others. 2025.
\newblock \href {https://arxiv.org/pdf/2401.14656} {Scientific large language models: A survey on biological \& chemical domains}.
\newblock \emph{ACM Computing Surveys}, 57(6):1--38.

\bibitem[{Zhang et~al.(2024{\natexlab{d}})Zhang, Chen, Jin, Wang, Ji, Wang, and Han}]{zhang2024comprehensive}
Yu~Zhang, Xiusi Chen, Bowen Jin, Sheng Wang, Shuiwang Ji, Wei Wang, and Jiawei Han. 2024{\natexlab{d}}.
\newblock \href {https://arxiv.org/pdf/2406.10833} {A comprehensive survey of scientific large language models and their applications in scientific discovery}.
\newblock \emph{arXiv preprint arXiv:2406.10833}.

\bibitem[{Zhao et~al.(2023)Zhao, Zhou, Li, Tang, Wang, Hou, Min, Zhang, Zhang, Dong et~al.}]{zhao2023survey}
Wayne~Xin Zhao, Kun Zhou, Junyi Li, Tianyi Tang, Xiaolei Wang, Yupeng Hou, Yingqian Min, Beichen Zhang, Junjie Zhang, Zican Dong, and 1 others. 2023.
\newblock \href {https://arxiv.org/pdf/2303.18223} {A survey of large language models}.
\newblock \emph{arXiv preprint arXiv:2303.18223}.

\bibitem[{Zheng et~al.(2024)Zheng, Koh, Yang, Li, May, Webb, Pan, and Church}]{zheng2024large}
Yizhen Zheng, Huan~Yee Koh, Maddie Yang, Li~Li, Lauren~T May, Geoffrey~I Webb, Shirui Pan, and George Church. 2024.
\newblock \href {https://arxiv.org/pdf/2409.04481} {Large language models in drug discovery and development: From disease mechanisms to clinical trials}.
\newblock \emph{arXiv preprint arXiv:2409.04481}.

\end{thebibliography}

\appendix
\clearpage

\begin{figure*}[ht]
\centering  
\includegraphics[height=2.1cm]{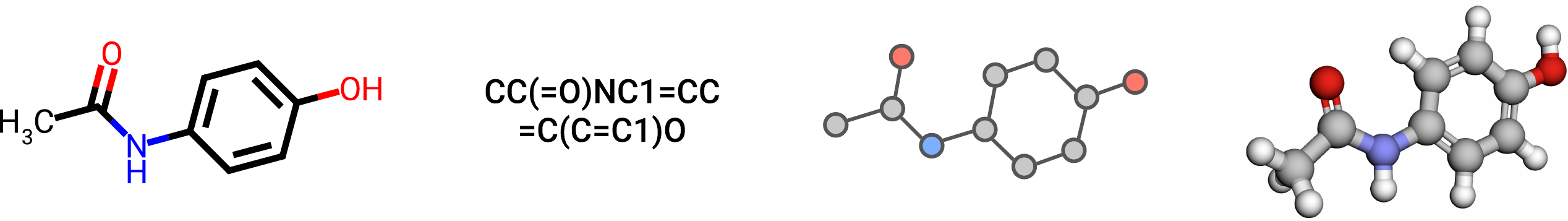}
\caption{\textbf{Illustration of an example molecule and its representation in different data modalities.}  From left to right following the 2D chemical structure diagram: its 1D SMILES string representation, a simplified 2D graph view, and its 3D ball-and-stick model.}
\label{fig: data_modality}  
\vspace{-0.3cm}
\end{figure*}

\section{Data Modalities for Molecular LLMs:}
\label{appendix:data_modality}

LLMs used for molecular generation and optimization interface with structured molecular data in various modalities. Each modality offers distinct structural or physicochemical information. As shown in Fig.~\ref{fig: data_modality}, commonly used molecular representations can be categorized into the following three formats:
\begin{itemize}[leftmargin=*, itemsep=-1pt, topsep=0.5pt]
\item \textbf{1D Sequence Representations ($S$)}: These are linear string encodings of molecular structures. Common formats include \textit{SMILES} (Simplified Molecular Input Line Entry System)~\citep{weininger1988smiles} and \textit{SELFIES} (Self-Referencing Embedded Strings)~\citep{krenn2020self}. These representations are well-suited for LLMs due to their compatibility with token-based language modeling. 
Another format used in certain settings is the IUPAC nomenclature~\citep{iupac2013nomenclature}, which provides systematic names for molecules and is employed as an alternative or auxiliary textual representation in language modeling frameworks.
\item \textbf{2D Graph Representations ($G$)}: A molecule is represented as a graph $G = (V, E)$, where nodes $v \in V$ correspond to atoms and edges $e \in E$ correspond to chemical bonds. Node and edge features may encode atom types, bond orders, aromaticity, and other topological attributes. While not directly token-based, 2D graphs can be integrated via hybrid models that combine language and graph encoders, or serialized (e.g., via adjacency lists or graph traversal sequences) to interface with LLMs.
\item \textbf{3D Geometric Representations ($X$)}: These representations capture atomic coordinates in three-dimensional space. Formally, $X = \{(a_i, \vec{r}_i)\}_{i=1}^{N}$, where $a_i$ denotes the atomic species and $\vec{r}_i \in \mathbb{R}^3$ specifies the Cartesian coordinates of atom $i$. 3D information is essential for modeling stereochemistry, conformational preferences, and interaction potentials. Incorporating 3D data into LLMs typically requires transforming it into a sequence-compatible format or using auxiliary models to predict or refine 3D structures.
\end{itemize}

\section{Datasets}
\label{appendix: datasets}
Datasets are crucial resources for advancing LLM-centric molecule design, serving extensively in both the training and evaluation phases of model development. Table~\ref{tab:datasets_summary} provides a comprehensive summary of commonly utilized molecule datasets, detailing their key features. For each dataset listed, the table specifies its \textbf{Last Update} year, approximate \textbf{Scale} (number of entries), whether it includes natural language \textbf{Instruction} components, and its suitability for \textbf{Pretraining} LLMs or as a \textbf{Benchmark} for evaluation. Furthermore, the table indicates the types of \textbf{Molecule Representations} available within each dataset, such as SMILES, IUPAC names, ready-to-dock formats (\textbf{Dock}), graph structures (\textbf{Graph}), 3D coordinates (\textbf{3D}), or formal chemical ontologies (\textbf{Ontology}). Finally, it highlights whether a dataset supports \textbf{Generation} or \textbf{Optimization} tasks, lists \textbf{Other Tasks} it is commonly used for (e.g., property prediction, translation), and provides a \textbf{Link} to access the resource. 

The subsequent subsections categorize these datasets based on their primary application focus, aligning with the classification used in Section~\ref{sec:benchmarking_evaluation} of the main text.

\subsection{Pretraining-Only Datasets}
Pretraining-only datasets typically contain diverse molecular structures and associated property information, designed to support broad generalization capabilities when pretraining LLMs for downstream tasks. These datasets generally do not include explicit natural language instructions or task-specific labels for direct supervised learning of specific generation or optimization objectives.
\begin{itemize}[leftmargin=*, itemsep=-1pt, topsep=0.5pt]
    \item \textbf{ZINC:} ZINC~\citep{irwin2012zinc} is a public and comprehensive database containing over 20 million commercially available molecules presented in biologically relevant representations. These molecules can be downloaded in popular ready-to-dock formats and various subsets, making ZINC widely used for distribution learning-based and goal-oriented molecule generation tasks.
    \item \textbf{PubChem:} PubChem~\citep{kim2016pubchem, kim2019pubchem, kim2025pubchem} serves as a vast public chemical information repository, holding over 750 million records. It covers a wide array of data, including chemical structures, identifiers, bioactivity outcomes, genes, proteins, and patents, and is organized into three interlinked databases: Substance (contributed chemical information), Compound (standardized unique structures), and BioAssay (biological experiment details).
    \item \textbf{ChemData:} ChemData~\citep{zhang2024chemllm} is a large-scale dataset specifically curated for fine-tuning chemical LLMs, containing 7 million instruction query-response pairs. Derived from various online structural datasets like PubChem and ChEMBL, it encompasses a broad range of chemical domain knowledge and is frequently used for tasks in molecule understanding, chemical process reasoning, and other domain-specific applications.
    \item \textbf{Mol-Instructions:} Mol-Instructions~\citep{fang2023mol} is a large-scale, diverse, and high-quality dataset designed for the biomolecular domain, featuring over 2 million carefully curated biomolecular instructions. It is structured around three core components: molecule-oriented instructions (148.4K across six tasks focusing on properties, reactions, and design), protein-oriented instructions (505K samples across five task categories related to protein structure, function, and design), and biomolecular text instructions (53K for bioinformatics and chemoinformatics NLP tasks like information extraction and question answering).
    \item \textbf{MuMOInstruct:} MuMOInstruct~\citep{dey2025mathtt} is presented as the first high-quality instruction-tuning dataset focused on complex, multi-property molecular optimization tasks. Unlike datasets such as MolOpt-Instruction~\citep{ye2025drugassist} that primarily target single- or dual-property tasks, MuMOInstruct emphasizes tasks involving at least three properties, facilitating the evaluation of LLMs in both in-domain and out-of-domain settings.
\end{itemize}

\begin{table*}[t!]
  \tiny
  \centering
  \caption{Summary of commonly used molecule datasets and their features. \textbf{Dock} denotes the "ready-to-dock" format; \textbf{Ontology} denotes the structured representation of the molecule; \textbf{Captioning} denotes molecule captioning task; \textbf{Docking} denotes molecule docking (a way to find correct molecule binds for proteins); \textbf{Translation} denotes the translation from textual knowledge to molecular features; \textbf{Conversion} denotes the translation between different representations of a molecule's identity; \textbf{Prediction} denotes property prediction, forward reaction prediction and retrosynthesis tasks; \textbf{QM} denotes hybrid quantum mechanics.}
  \renewcommand{\arraystretch}{1.4}
  \setlength\tabcolsep{0.9pt}
  \begin{tabular}{@{} 
      c | c | c | c | c  c | 
      c  c  c  c  c 
      c | c  c  c | c  
    @{}}
    \bottomrule
    \multirow{2}{*}{\textbf{Datasets}} & 
    \multirow{2}{*}{\textbf{\makecell{Last\\Update}}} & 
    \multirow{2}{*}{\textbf{Scale}} & 
    \multirow{2}{*}{\textbf{\makecell{Instruc-\\tion}}} & 
    \multirow{2}{*}{\textbf{\makecell{Pretrain-\\ing}}} & 
    \multirow{2}{*}{\textbf{\makecell{Bench\\-mark}}} & 
    \multicolumn{6}{c|}{\textbf{Molecule Representations}} & 
    \multirow{2}{*}{\textbf{\makecell{Genera-\\tion}}} & 
    \multirow{2}{*}{\textbf{\makecell{Optimi-\\zation}}} & 
    \multirow{2}{*}{\textbf{Other Tasks}} & 
    \multirow{2}{*}{\textbf{Link}} \\
    & & & & & & 
    \textbf{SMILES} & \textbf{IUPAC} & \textbf{Dock} & \textbf{Graph} & \textbf{3D} & \textbf{Ontology} & 
    & & &  \\
    \midrule
    \makecell{PubChem\\\citep{kim2016pubchem, kim2019pubchem, kim2025pubchem}} & 2025 & 119M & \xmark & \cmark & \xmark 
  & \cmark & \cmark & \xmark & \cmark & \cmark & \cmark & \cmark & \xmark & Property Prediction \& Biology Domain & \href{https://pubchem.ncbi.nlm.nih.gov/docs/downloads}{Link} \\
\makecell{ChEMBL\\\citep{gaulton2012chembl}} & 2024 & >20M & \xmark & \cmark & \cmark 
  & \cmark & \cmark & \xmark & \cmark & \xmark & \xmark & \cmark & \cmark & Prediction \& ML Benchmark & \href{https://ftp.ebi.ac.uk/pub/databases/chembl/ChEMBLdb/releases/chembl\_35/}{Link} \\
\makecell{CrossDocked2020\\\citep{francoeur2020three}} & 2024 & 22.5M & \xmark & \cmark & \cmark 
  & \cmark & \xmark & \cmark & \xmark & \cmark & \xmark & \xmark & \cmark & Docking Datasets & \href{https://github.com/gnina/models/tree/master/data/CrossDocked2020}{Link} \\
\makecell{ZINC\\\citep{irwin2012zinc}} & 2023 & >980M & \xmark & \cmark & \xmark 
  & \cmark & \cmark & \cmark & \cmark & \cmark & \xmark & \cmark & \cmark & Ligand Discovery & \href{https://wiki.docking.org/index.php?title=ZINC15:Getting\_started\#Downloading\_SMILES}{Link} \\
\makecell{Dockstring\\\citep{garcia2022dockstring}} & 2022 & >260k & \xmark & \cmark & \cmark 
  & \cmark & \xmark & \cmark & \cmark & \cmark & \xmark & \cmark & \cmark & Virtual Screening & \href{https://figshare.com/articles/dataset/dockstring\_dataset/16511577}{Link} \\
\makecell{ChEBI-20\\\citep{edwards2021text2mol}} & 2021 & 33k & \xmark & \cmark & \cmark 
  & \cmark & \cmark & \xmark & \cmark & \xmark & \cmark & \cmark & \xmark & Translation \& Classification \& Captioning & \href{https://github.com/cnedwards/text2mol/tree/master/data}{Link} \\
\makecell{OGBG-MolHIV\\\citep{hu2020open}} & 2020 & $\sim$41k & \xmark & \cmark & \cmark 
  & \cmark & \xmark & \xmark & \cmark & \xmark & \xmark & \cmark & \xmark & Graph Property Prediction & \href{https://ogb.stanford.edu/docs/graphprop/\#ogbg-mol}{Link} \\
\makecell{MOSES\\\citep{polykovskiy2020molecular}} & 2020 & $\sim$1.9M & \xmark & \xmark & \cmark 
  & \cmark & \xmark & \xmark & \xmark & \xmark & \xmark & \cmark & \xmark & De novo Design & \href{https://github.com/molecularsets/moses}{Link} \\
\makecell{MoleculeNet\\\citep{wu2018moleculenet}} & 2019 & 700k & \xmark & \xmark & \cmark 
  & \cmark & \xmark & \xmark & \cmark & \cmark & \xmark & \cmark & \cmark & ML Benchmark & \href{https://github.com/deepchem/deepchem}{Link} \\
\makecell{QM9\\\citep{pinheiro2020machine}} & 2014 & 134k & \xmark & \cmark & \cmark 
  & \cmark & \xmark & \xmark & \cmark & \cmark & \xmark & \cmark & \cmark & Hybrid QM/ML Modeling  & \href{https://figshare.com/collections/Quantum\_chemistry\_structures\_and\_properties\_of\_134\_kilo\_molecules/978904/5}{Link} \\
\makecell{TOMG-Bench\\\citep{li2024tomg}} & 2025 & 5k & \cmark & \xmark & \cmark 
  & \cmark & \xmark & \xmark & \cmark & \cmark & \cmark & \cmark & \cmark & Molecule Editing & \href{https://github.com/phenixace/TOMG-Bench}{Link} \\
\makecell{MuMOInstruct\\\citep{dey2025mathtt}} & 2025 & 873k & \cmark & \cmark & \xmark 
  & \cmark & \xmark & \xmark & \xmark & \xmark & \xmark & \xmark & \cmark & — & \href{https://huggingface.co/datasets/NingLab/MuMOInstruct}{Link} \\
\makecell{ChemData\\\citep{zhang2024chemllm}} & 2024 & 7M & \cmark & \cmark & \xmark 
  & \cmark & \xmark & \cmark & \xmark & \xmark & \xmark & \cmark & \cmark & Conversion \& Prediction \& Reaction & \href{https://huggingface.co/datasets/AI4Chem/ChemData700K}{Link} \\
\makecell{ChemBench\\\citep{mirza2404large}} & 2024 & 4k & \cmark & \xmark & \cmark 
  & \cmark & \xmark & \xmark & \xmark & \xmark & \xmark & \cmark & \cmark & Reaction Benchmark \& Virtual Screening & \href{https://huggingface.co/datasets/jablonkagroup/ChemBench}{Link} \\
\makecell{Mol-Instructions\\\citep{fang2023mol}} & 2024 & 2M & \cmark & \cmark & \xmark 
  & \cmark & \xmark & \xmark & \xmark & \xmark & \xmark & \cmark & \cmark & Translation, Retrosynthesis & \href{https://huggingface.co/collections/zjunlp/mol-instructions-662e0b9435ab6df9593e8ea0}{Link} \\
\makecell{MolOpt-Instructions\\\citep{ye2025drugassist}} & 2024 & 1M & \cmark & \cmark & \cmark 
  & \cmark & \xmark & \xmark & \xmark & \xmark & \xmark & \xmark & \cmark & — & \href{https://huggingface.co/datasets/blazerye/MolOpt-Instructions}{Link} \\
\makecell{L\,+M-24\\\citep{edwards2024l+}} & 2024 & 148k & \cmark & \cmark & \cmark 
  & \cmark & \xmark & \xmark & \cmark & \xmark & \xmark & \cmark & \xmark & Captioning & \href{https://github.com/language-plus-molecules/LPM-24-Dataset}{Link} \\
\makecell{SMolInstruct\\\citep{yu2024advancing}} & 2024 & 3.3M & \cmark & \cmark & \cmark 
  & \cmark & \xmark & \xmark & \xmark & \xmark & \xmark & \cmark & \xmark & Captioning \& Prediction & \href{https://huggingface.co/datasets/osunlp/SMolInstruct/tree/main}{Link} \\
    \bottomrule
  \end{tabular}
  \label{tab:datasets_summary}
\end{table*}

\subsection{Benchmark-Only Datasets} 
Benchmark-only datasets are specifically curated for the evaluation of models, particularly in generative molecular tasks. These datasets often feature structured input-output pairs, such as instruction-molecule pairings, and are typically smaller in scale, manually verified, and tailored to specific evaluative purposes.
\begin{itemize}[leftmargin=*, itemsep=-1pt, topsep=0.5pt]
    \item \textbf{MoleculeNet:} A large-scale benchmark compendium, MoleculeNet~\citep{wu2018moleculenet} is derived from multiple public databases. It comprises 17 curated datasets with over 700,000 compounds, represented textually (e.g., SMILES) and in 3D formats. Covering a wide array of properties categorized into quantum mechanics, physical chemistry, biophysics, and physiology, it serves as a standard for evaluating molecular property prediction models.
    \item \textbf{ChemBench:} ChemBench~\citep{mirza2404large} offers a comprehensive framework for benchmarking the chemical knowledge and reasoning abilities of LLMs. It consists of thousands of manually curated question-answer pairs from diverse sources, focusing on three core aspects: Calculation, Reasoning, and Knowledge.
    \item \textbf{TOMG-Bench:} As the first benchmark dedicated to the open-domain molecule generation capabilities of LLMs, TOMG-Bench (Text-based Open Molecule Generation Benchmark)~\citep{li2024tomg} contains 45,000 samples. It is structured around three primary tasks: molecule editing (MolEdit), molecule optimization (MolOpt), and customized molecule generation (MolCustom).
    \item \textbf{MOSES:} MOSES (Molecular Sets) \citep{polykovskiy2020molecular} is a task-specific resource designed for both training and benchmarking molecule generation models in drug discovery. Containing approximately 1.9 million molecules in SMILES format derived from the ZINC Clean Leads dataset, it also furnishes training, testing, and scaffold-split subsets, along with built-in evaluation metrics.
\end{itemize}

\subsection{Datasets for Pretraining \& Benchmark Applications} 
A distinct category of datasets offers the flexibility to be used for both pretraining LLMs and for subsequent benchmarking. These resources often combine substantial scale with features amenable to diverse evaluation scenarios.
\begin{itemize}[leftmargin=*, itemsep=-1pt, topsep=0.5pt]
    \item \textbf{ChEMBL:} ChEMBL~\citep{gaulton2012chembl} is a manually curated, open-access database focusing on drug-like bioactive molecules. It houses 5.4 million bioactivity measurements for over 1 million compounds and 5,200 protein targets, effectively integrating chemical, bioactivity, and genomic data to support drug discovery and the translation of genomic insights into therapeutics.
    \item \textbf{ChEBI-20:} ChEBI-20~\citep{edwards2021text2mol}, derived from the ChEBI database, is a freely available, manually curated dictionary of molecular entities concentrated on small chemical compounds. It includes over 20,000 molecules represented by SMILES strings, natural language descriptions, and ontology terms, widely employed in molecule generation and instruction-based tasks requiring chemical understanding.
    \item \textbf{CrossDocked2020:} CrossDocked2020~\citep{francoeur2020three} is a large-scale dataset specifically geared towards structure-based drug design (SBDD). It features over 22 million 3D docked poses of protein-ligand pairs, making it a valuable resource for tasks like pocket-conditioned 3D molecule generation.
    \item \textbf{Dockstring:} Dockstring~\citep{garcia2022dockstring} provides a large-scale, well-curated dataset for molecular docking. It encompasses an extensive collection of docking scores and poses for more than 260,000 ligands against 58 medically relevant targets, and includes pharmaceutically relevant benchmark tasks such as virtual screening and the \textit{de novo} design of selective kinase inhibitors.
    \item \textbf{QM9:} QM9(The Quantum Mechanics 9) dataset~\citep{pinheiro2020machine} is a public quantum chemistry resource containing approximately 134,000 small organic molecules (composed of H, C, N, O, F; up to nine non-hydrogen atoms). It provides SMILES representations, 3D geometries, and quantum chemical properties, widely utilized for training and evaluating molecular property prediction models.
    \item \textbf{SMolInstruct:} SMolInstruct~\citep{yu2024advancing} is a large-scale, comprehensive, and high-quality dataset for instruction tuning LLMs in chemistry. It consists of 3.3 million language-molecule pairs and 1.6 million distinct molecules, covering four types of molecular representations and 14 different tasks, with molecules represented in SMILES or SELFIES format.
    \item \textbf{OGBG-MolHIV:} OGBG-MolHIV~\citep{hu2020open}, part of the Open Graph Benchmark, is an open-access, task-specific dataset for binary molecular property prediction, specifically for classifying HIV inhibition. It contains 41,127 unique molecules in graph format, where nodes (atoms) have 9 numerical features and edges (bonds) have 3-dimensional features (type, stereochemistry, conjugation). It is derived from MoleculeNet and preprocessed using RDKit.
    \item \textbf{MolOpt-Instructions:} MolOpt-Instructions~\citep{ye2025drugassist} is an instruction-based dataset tailored for molecule optimization, containing over 1 million molecule-molecule pairs. It was constructed by selecting molecules from ZINC and using MMPDB to generate and filter for highly similar pairs, covering six molecular properties including solubility, BBBP, and hERG inhibition.
    \item \textbf{L+M-24:} L+M-24 (Language + Molecules 24 Tasks)~\citep{edwards2024l+} is a large-scale, multi-task instruction dataset designed to leverage the benefits of natural language (compositionality, functionality, abstraction) in molecule design. Derived from PubChem and other sources, it contains over 148,000 language-molecule pairs spanning 24 distinct molecule design tasks across various application domains.
\end{itemize}

\begin{figure}[t!]
\centering
\includegraphics[height=5.95cm]{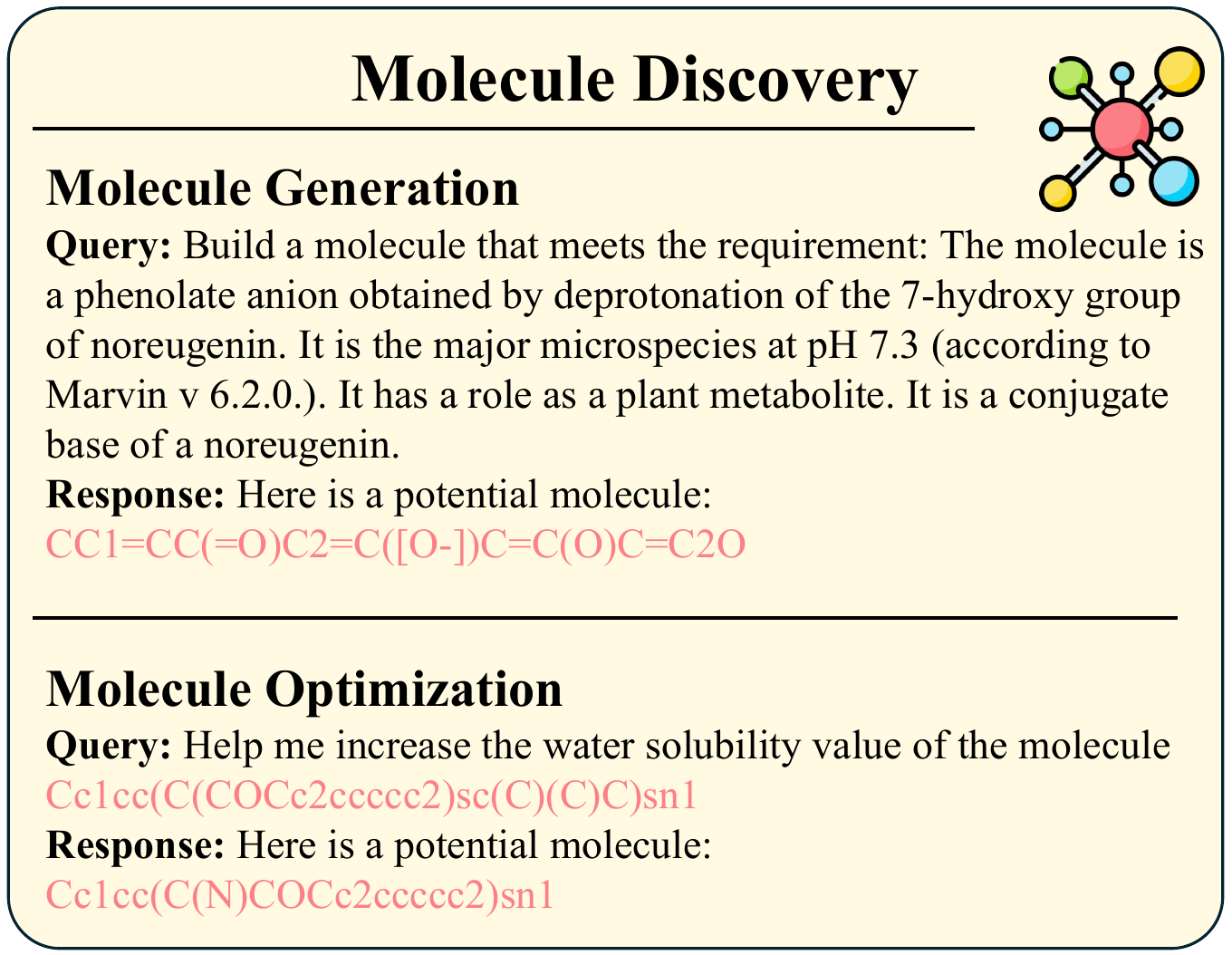}
\caption{Visualization of the Instruction dataset of molecule generation and optimization task.}
\label{fig: cap2mol}
\vspace{-0.3cm}
\end{figure}

\section{Evaluation Metrics}
\label{appendix:metrics} 

In the context of LLM-centric molecular generation and optimization, evaluation metrics are commonly grouped into categories based on molecular structure, physicochemical properties, and optimization success, each reflecting distinct aspects of molecular quality and model performance. This appendix details these metrics, aligning with the categorization presented in Fig.~\ref{fig:evaluation_taxonomy}. 

\subsection{Structure-Based Metrics}
Structure-based metrics are employed to assess the chemical plausibility, resemblance to reference compounds, and structural diversity of molecules generated or modified by LLMs. These metrics help ensure that the outputs are chemically meaningful and cover a sufficient breadth of the relevant chemical space.

\subsubsection{Validity \& Similarity}
Metrics for validity and similarity evaluate the extent to which generated molecules conform to chemical rules, match the structural features of reference molecules, and satisfy specified structural constraints. They are crucial for determining if the outputs are chemically sound and potentially useful for applications like drug discovery.
\begin{itemize}[leftmargin=*, itemsep=-1pt, topsep=0.5pt]
    \item \textbf{Validity Rate:} The validity rate~\citep{polykovskiy2020molecular} indicates the fraction of generated molecules that are chemically valid (e.g., parsable by RDKit) and often also considers uniqueness among valid structures. A high validity rate suggests that the LLM has effectively learned the underlying rules of molecular representation (e.g., SMILES grammar) and the context provided by textual descriptions.
    \item \textbf{EM (Exact Match):} Exact Match~\citep{rajpurkar2016squad} assesses whether a generated molecular sequence is identical to a target reference sequence. A higher EM rate signifies a stronger capability of the model to precisely replicate reference molecules when required.
    \item \textbf{BLEU (Bilingual Evaluation Understudy):} This score~\citep{papineni2002bleu}, originally from machine translation, measures the n-gram overlap between generated and reference molecular sequences. A higher BLEU score indicates greater similarity in token order and composition, reflecting better fidelity to the ground truth molecule's sequence.
    \item \textbf{Levenshtein Distance:} The Levenshtein distance~\citep{levenshtein1966binary} quantifies the dissimilarity between two strings by counting the minimum number of single-character edits (insertions, deletions, or substitutions) required to change one string into the other. A lower Levenshtein distance signifies higher similarity between the generated and reference molecular sequences.
    \item \textbf{FTS (Fingerprint Tanimoto Similarity):} FTS is a widely used metric for quantifying structural similarity based on molecular fingerprints, such as MACCS~\citep{durant2002reoptimization}, RDKit topological fingerprints (RDK)~\citep{landrum2013rdkit}, or Morgan fingerprints (circular fingerprints)~\citep{morgan1965generation}. A higher FTS score (typically ranging from 0 to 1) indicates a greater overlap in key substructures and chemical patterns between the generated and reference molecules.
    \item \textbf{FCD (Fréchet ChemNet Distance):} FCD \citep{preuer2018frechet} evaluates the dissimilarity between the distribution of features (derived from a pretrained chemical neural network) of generated molecules and a reference set (often ground truth molecules). Lower FCD values suggest that the generated molecules better capture the chemical diversity and property distribution of the reference set.
\end{itemize}

\subsubsection{Diversity \& Uniqueness}
Metrics related to diversity, uniqueness, and novelty assess an LLM's ability to produce a varied set of outputs. High performance in these areas can help prevent mode collapse and enhance the exploration of chemical space for discovering novel and relevant molecules.
\begin{itemize}[leftmargin=*, itemsep=-1pt, topsep=0.5pt]
    \item \textbf{Uniqueness:} Uniqueness quantifies the proportion of valid generated molecules that are distinct from each other within a given set. It reflects the model’s capacity to generate diverse structures rather than redundant outputs. This is often evaluated at different scales, such as Unique@1k (within the first 1,000 valid samples)~\citep{wang2023cmol} and Unique@10k (within 10,000 valid samples)~\citep{bagal2021molgpt}, to assess short-range and broader diversity, respectively.
    \item \textbf{Novelty Rate:} The novelty rate~\citep{brown2019guacamol} measures the fraction of valid and unique generated molecules that are not present in the training dataset. It serves as an indicator of the model's generalization ability and its potential to discover previously unseen chemical entities. A low novelty rate may suggest overfitting.
    \item \textbf{IntDiv (Internal Diversity) and NCircles:} These metrics further characterize the structural diversity within a set of generated molecules. 
    IntDiv~\citep{benhenda2017chemgan} calculates the average dissimilarity (1 minus Tanimoto similarity) between all pairs of molecules in the generated set, often using a power mean to adjust sensitivity:
    \[ \text{IntDiv}_p(S) = 1 - \left( \frac{1}{|S|^2} \sum_{s_i, s_j \in S} T(s_i, s_j)^p \right)^{\frac{1}{p}} \]
    where $T(s_i, s_j)$ is the Tanimoto similarity between molecules $s_i$ and $s_j$.
    NCircles~\citep{jang2024can} measures the size of the largest subset of generated molecules where no two molecules have a Tanimoto similarity exceeding a predefined threshold. A higher NCircles value indicates greater structural dissimilarity within the set.
\end{itemize}

\subsection{Property-Based Metrics}
Property-based metrics evaluate whether a designed or modified molecule satisfies specific physicochemical or biological property constraints, often crucial for assessing its potential utility, such as drug-likeness or target activity.

\subsubsection{Single-Property Evaluation Metrics}
In single-property evaluation, the primary goal is to assess the model's ability to generate or optimize molecules with respect to a specific molecular property, such as drug-likeness, solubility, or binding affinity.
\begin{itemize}[leftmargin=*, itemsep=-1pt, topsep=0.5pt]
    \item \textbf{LogP (Octanol-Water Partition Coefficient):} LogP~\citep{hansch1968linear} is the logarithm of a compound's partition coefficient between octanol and water, serving as a key indicator of molecular hydrophobicity and thus, often correlating with membrane permeability.
    \item \textbf{QED (Quantitative Estimate of Drug-likeness):} QED~\citep{bickerton2012quantifying} provides a heuristic score (ranging from 0 to 1) that integrates multiple physicochemical properties (e.g., molecular weight, LogP, number of hydrogen bond donors/acceptors) to estimate a compound's overall drug-likeness.
    \item \textbf{TPSA (Topological Polar Surface Area):} TPSA~\citep{ertl2000fast} quantifies the surface sum over all polar atoms in a molecule, reflecting its ability to form hydrogen bonds. It is often correlated with properties like intestinal absorption and blood-brain barrier penetration.
    \item \textbf{SA Score (Synthetic Accessibility Score):} The SA score~\citep{ertl2009estimation} estimates the ease of synthesizing a compound, typically on a scale from 1 (easy to synthesize) to 10 (very difficult to synthesize), based on fragment contributions and complexity penalties.
\end{itemize}

\subsubsection{Multi-Property Evaluation Metrics}
Multi-property evaluation assesses a model's performance in satisfying multiple, often competing, objectives simultaneously. This is critical in real-world scenarios where a balance of several properties is required.
\begin{itemize}[leftmargin=*, itemsep=-1pt, topsep=0.5pt]
    \item \textbf{Composite Score:} A composite score~\citep{jin2020multi} aggregates multiple individual property scores into a single scalar objective, often through a weighted sum or other combination rules. This allows optimization frameworks (e.g., evolutionary algorithms, reinforcement learning) to be guided by a unified fitness metric. The weights can be adjusted to reflect task-specific priorities among properties like LogP, QED, and synthetic accessibility.
    \item \textbf{Pareto Optimality:} In multi-objective optimization, a solution is considered Pareto optimal~\citep{pareto1919manuale} if none of its objective function values can be improved without degrading at least one of the other objective values. The set of all Pareto optimal solutions forms the Pareto front, which is used to visualize and analyze trade-offs between conflicting objectives.
    \item \textbf{Success Rate under Constraints:} This metric~\citep{jin2020multi} quantifies the proportion of generated or modified molecules that successfully meet or exceed predefined target thresholds across all specified properties. A common instantiation is the multi-property hit ratio, where a molecule is deemed successful only if all targeted property improvements satisfy their respective criteria.
\end{itemize}

\section{Method Summary}
\label{sec:method_summary}

This section provides a consolidated overview of representative LLM-based methods for molecular discovery, as detailed in Table~\ref{tab:summary_tab}. The table organizes these approaches primarily by the two core task categories central to this survey: molecule generation and molecule optimization. Within each task, methods are further sub-categorized by their primary learning Strategy (referred to as "Category" and "Technique" in the table), encompassing approaches without LLM tuning (such as zero-shot prompting and in-context learning) and those with LLM tuning (supervised fine-tuning and preference tuning). 

Table~\ref{tab:summary_tab} details several key aspects for each listed \textbf{Method}:
\begin{itemize}[leftmargin=*, itemsep=0pt, topsep=0.2em]
    \item \textbf{Venue}: The publication venue or preprint archive where the method was reported.
    \item \textbf{Input Type}: Specifies the primary format of molecular data and instructions provided to the LLM (e.g., SMILES strings, textual instructions, few-shot examples, or multi-modal inputs like graphs).
    \item \textbf{Base Model}: Indicates the foundational LLM architecture (e.g., GPT-4, LLaMA variants, Mistral) upon which the method is built or applied.
    \item \textbf{Dataset}: Lists the key molecular corpora or benchmarks used for training the model (if applicable) or for its evaluation in the context of the reported work.
    \item \textbf{Repository}: Provides a link to the public code or resource repository, if available.
\end{itemize}
This structured presentation aims to offer a clear comparative landscape of the current methodologies in the field.

\begin{table*}[ht]
\setlength\tabcolsep{2.pt}
\centering
\caption{Summary of LLM-based methods for molecule generation and optimization. Each row corresponds to a method, organized by \textbf{Task} (generation or optimization), and \textbf{Technique}. \textbf{Input Type} denotes the molecular data format provided to the model. \textbf{Base Model} denotes the large language model architecture used. \textbf{Dataset} denotes the molecular corpus or benchmark used for training or evaluation.}
\renewcommand{\arraystretch}{1.4}
\scalebox{0.6}{
\begin{tabular}{c|c|ccccccc}
\bottomrule
  &  &  &  &  &  &  &  \\  
 \multirow{-2}{*}{\textbf{Task}} & \multirow{-2}{*}{\textbf{Category}} & \multirow{-2}{*}{\textbf{Technique}} & \multirow{-2}{*}{\textbf{Method}} & \multirow{-2}{*}{\textbf{Venue}} & \multirow{-2}{*}{\makecell{\textbf{Input} \\ \textbf{Type}}} & \multirow{-2}{*}{\makecell{\textbf{Base} \\ \textbf{Model}}} & \multirow{-2}{*}{\textbf{Dataset}} & \multirow{-2}{*}{\textbf{Repository}}  \\ \toprule
\multirow{30}{*}{\rotatebox{90}{\makecell{Molecule \\ Generation}}} &
\multirow{4}{*}{\makecell{w/o\\Tuning}} &
\multirow{4}{*}{\makecell{ICL}}
& \makecell{LLM4GraphGen\\\citep{yao2024exploring}} & Arxiv & \makecell{Instruction + \\ Few shot} & GPT-4             & OGBG-MolHIV    & \href{https://github.com/SitaoLuan/LLM4Graph}{Link}     \\ \cline{4-9}
 & & & \makecell{MolReGPT\\\citep{li2024empowering}}  & TKDE & \makecell{Instruction + \\ Few shot} & \makecell{GPT-3.5-turbo/ \\ GPT-4}            & ChEBI-20    &  \href{https://github.com/phenixace/MolReGPT}{Link}      \\ \cline{4-9}
 & & & \makecell{FrontierX\\\citep{srinivas2024cross}} & Arxiv & Instruction            & GPT-3.5           & ChEBI-20      &   N/A   \\ \cline{2-9}
&\multirow{22}{*}{\makecell{w/\\Tuning}} & \multirow{15}{*}{\makecell{SFT}}
& \makecell{Mol-instructions\\\citep{fang2023mol}} & ICLR & Instruction            & LLaMA-7B          & Mol-Instructions  & \href{https://github.com/zjunlp/Mol-Instructions}{Link} \\ \cline{4-9}
 & & & \makecell{LlaSMol\\\citep{yu2024llasmol}}    &   COLM   & Instruction            & \makecell{Galactica 6.7B/\\LLaMA-2-7B/\\ Mistral-7B}         & SMolInstruct     & \href{https://osu-nlp-group.github.io/LLM4Chem/}{Link}  \\ \cline{4-9}
 & & & \makecell{ChemLLM\\\citep{zhang2024chemllm}}  &    Arxiv  & Instruction            & \makecell{InternLM2- \\ 7B-Base} & ChemData   & N/A        \\ \cline{4-9}
 & & & \makecell{ICMA\\\citep{li2024large}}        &  TKDE    & \makecell{Instruction + \\ Few shot} & Mistral-7B        & \makecell{PubChem \& \\ ChEBI-20} & N/A \\ \cline{4-9}
 & & & MolReFlect~\citep{li2024molreflect} & Arxiv  & \makecell{Instruction + \\ Few shot} & Mistral-7B        & ChEBI-20    &     \href{https://github.com/vllm-project/vllm}{Link}     \\ \cline{4-9}
 & & & \makecell{ChatMol\\\citep{fan2025chatmol}}     & Arxiv   & Instruction            & LLaMA-3-8B         & ZINC        &    \href{https://github.com/ChatMol/ChatMol}{Link}     \\ \cline{4-9}
 & & & \makecell{PEIT-LLM\\\citep{lin2025prop}}      &  Arxiv  & Instruction            & \makecell{LLaMA-3.1-8B/ \\ Qwen2.5-7B}       & ChEBI-20     &   \href{https://github.com/}{Link}    \\ \cline{4-9}
 & & & \makecell{NatureLM \\\citep{xia2025naturelm}}      &  Arxiv  & \makecell{SMILES + \\ Instruction}            & NatureLM-8B       & \makecell{ChEMBL \& \\ MoleculeNet}      &   \href{https://naturelm.github.io/}{Link}    \\ \cline{4-9}
 & & & \makecell{SynLlama
 \\\citep{sun2025synllama}}      &  Arxiv  & Instruction            & \makecell{LLaMA-3.1-8B / \\LLaMA-3.2-1B}       & ChEMBL     &   \href{https://github.com/THGLab/SynLlama}{Link}    \\ \cline{4-9}
 & & & \makecell{TOMG-Bench
 \\\citep{li2024tomg}}      &  Arxiv  & Instruction   & LLaMa-3.1-8B    & TOMG-Bench   &   N/A    \\ \cline{4-9}
  & & & \makecell{UniMoT\\\citep{zhang2024unimot}}    &     Arxiv  & Instruction            & LLaMA-2-7B        & Mol-Instructions & \href{https://uni-mot.github.io/}{Link}\\ \cline{3-9}
& & \multirow{8}{*}{\makecell{Preference \\ Tuning}}
& \makecell{Div-SFT\\\citep{jang2024can}}     &    Arxiv   & Instruction            & LLaMA-7B          & ChEBI-20     &  N/A    \\ \cline{4-9}
 & & & \makecell{Mol-MOE\\\citep{calanzone2025mol}}     & Arxiv  & Instruction            & LLaMA-3.2-1B      & \makecell{ChEMBL \& \\ ZINC \& \\ MOSES} & \href{https://github.com/ddidacus/mol-moe}{Link} \\ \cline{4-9}
 & & & \makecell{SmileyLLama\\\citep{cavanagh2024smileyllama}} & NeurIPS Workshop & Instruction      & LLaMA-3.1-8B      & ChEMBL       &  N/A      \\ \cline{4-9}
 & & & \makecell{ALMol\\\citep{gkoumas2024almol}}    &   ACL Workshop   & Instruction            & Meditron-7B       & L+M-24      &    N/A    \\ \cline{4-9}
 & & & \makecell{Less for More\\\citep{gkoumas2024less}}  & Arxiv & Instruction            & Meditron-7B       & L+M-24      &   N/A      \\ \cline{4-9}
  & & & \makecell{Mol-LLM\\\citep{lee2025mol}}      &   Arxiv   & Instruction            & Mistral-7B        & ChEBI-20      &   N/A     \\ 
\midrule 
\multirow{25}{*}{\rotatebox{90}{\makecell{Molecule Optimization}}} &
\multirow{12}{*}{{\makecell{w/o\\Tuning}}} &
\multirow{3}{*}{\makecell{Zero-Shot \\ Prompting}}
& \makecell{LLM-MDE\\\citep{bhattacharya2024large}} & JCIM & \makecell{SMILES + \\ Instruction} & Claude 3 Opus & ZINC & N/A \\ \cline{4-9}
& & & \makecell{MOLLEO\\\citep{wang2025efficient}} & ICLR & \makecell{SMILES + \\ Instruction} & GPT-4  & ZINC & \href{https://github.com/zoom-wang112358/MOLLEO}{Link}\\ \cline{3-9}
& & \multirow{10}{*}{\makecell{ICL}}
& \makecell{CIDD\\\citep{gao2025pushing}} & Arxiv & \makecell{SMILES + \\ Interaction report} & GPT-4o & CrossDocked2020 & N/A \\ \cline{4-9}
& & & \makecell{LLM-EO\\\citep{lu2024generative}} & Arxiv &  \makecell{SMILES + \\ Ligands Pool} & \makecell{Claude 3.5 Sonnet / \\ OpenAI o1-preview} & TMC dataset & \href{https://github.com/deepprinciple/llmeo}{Link} \\ \cline{4-9}
& & & \makecell{MOLLM\\\citep{ran2025multi}} & Arxiv & \makecell{SMILES + \\ Instruction} & GPT-4o & ZINC & N/A \\ \cline{4-9}
& & & \makecell{ChatDrug\\\citep{liu2024conversational}} & ICLR & \makecell{SMILES + \\ Instruction} & \makecell{Galactica / \\ LLaMA-2 / \\ ChatGPT} & ZINC & \href{https://github.com/chao1224/ChatDrug}{Link}\\ \cline{4-9}
& & & \makecell{Re$^2$DF\\\citep{le2024utilizing}} & Arxiv & \makecell{SMILES + \\ Instruction} & \makecell{LLaMA-3.1-8B/ \\ LLaMA-3.1-70B} & ZINC &  \href{https://github.com/lhkhiem28/Re2DF}{Link}\\ \cline{4-9}
& & & \makecell{BOPRO\\\citep{agarwalsearching}}  & ICLR & \makecell{SMILES + \\ Instruction} & Mistral-Large-Instruct-2407 & Dockstring & \href{https://github.com/amazon-science/BOPRO-ICLR-2025}{Link} \\ \cline{2-9}
& \multirow{12}{*}{{\makecell{w/\\Tuning}}} & 
\multirow{10}{*}{\makecell{SFT}}
& \makecell{MultiMol\\\citep{yu2025collaborative}} & Arxiv & \makecell{SMILES + \\ Instruction} & \makecell{Qwen2.5-7B / \\ LLaMA-3.1-8B / \\ Galactica 6.7B} & PubChem & \href{https://github.com/jiajunyu1999/LLM4Drug}{Link} \\ \cline{4-9}
& & & \makecell{DrugAssist\\\citep{ye2025drugassist}} & \makecell{Brief Bioinform} & \makecell{SMILES + \\ Instruction} & LLaMA-2-7B-Chat & MolOpt-Instructions & \href{https://github.com/blazerye/DrugAssist}{Link}\\ \cline{4-9}
& & & \makecell{GeLLM$^3$O\\\citep{dey2025mathtt}}  & Arxiv & \makecell{SMILES + \\ Instruction} & \makecell{Mistral-7B-Instruct / \\ LLaMA-3.1-8B-Instruct} & MuMOInstruct & \href{https://github.com/ninglab/GeLLMO}{Link} \\ \cline{4-9}
& & & \makecell{DrugLLM\\\citep{liu2024drugllm}}  & Arxiv & \makecell{Group-based \\ Molecular Representation} & LLaMA-2-7B & \makecell{ZINC \& \\ ChEMBL} & N/A \\ \cline{4-9}
& & & \makecell{TOMG-Bench
 \\\citep{li2024tomg}}      &  Arxiv  & Instruction   & LLaMa-3.1-8B    & TOMG-Bench   &   N/A    \\ \cline{4-9}
& & & \makecell{LLM-Enhanced GA\\\citep{bedrosian2024small}} & NeurIPS Workshop  & JSON Objects & \makecell{Chemma / \\ Chemlactica} & PubChem & \href{https://github.com/yerevann/chemlactica}{Link} \\ \cline{4-9}
& & & \makecell{Molx-Enhanced LLM\\\citep{le2024molx}} & Arxiv & \makecell{SMILES + \\ Graph + \\ Instruction}  & LLaMA-2-7B & PubChem & N/A \\ \cline{3-9}
& & \multirow{1}{*}{\makecell{Preference\\ Tuning}}
& 
\makecell{NatureLM\\\citep{xia2025naturelm}} & Arxiv & \makecell{SMILES + \\ Instruction} & NatureLM-8B & \makecell{ChEMBL \& \\ MoleculeNet}  & N/A \\
\bottomrule
\end{tabular}}
\label{tab:summary_tab}
\end{table*}




\end{document}